# Wikipedia-based Semantic Interpretation for Natural Language Processing

**Evgeniy Gabrilovich**                                    GABR@YAHOO-INC.COM
**Shaul Markovitch**                                       SHAULM@CS.TECHNION.AC.IL
*Department of Computer Science*
*Technion—Israel Institute of Technology*
*Technion City, 32000 Haifa, Israel*

## Abstract

Adequate representation of natural language semantics requires access to vast amounts of common sense and domain-specific world knowledge. Prior work in the field was based on purely statistical techniques that did not make use of background knowledge, on limited lexicographic knowledge bases such as WordNet, or on huge manual efforts such as the CYC project. Here we propose a novel method, called Explicit Semantic Analysis (ESA), for fine-grained semantic interpretation of unrestricted natural language texts. Our method represents meaning in a high-dimensional space of concepts derived from Wikipedia, the largest encyclopedia in existence. We explicitly represent the meaning of any text in terms of Wikipedia-based concepts. We evaluate the effectiveness of our method on text categorization and on computing the degree of semantic relatedness between fragments of natural language text. Using ESA results in significant improvements over the previous state of the art in both tasks. Importantly, due to the use of natural concepts, the ESA model is easy to explain to human users.

## 1. Introduction

Recent proliferation of the World Wide Web, and common availability of inexpensive storage media to accumulate over time enormous amounts of digital data, have contributed to the importance of intelligent access to this data. It is the sheer amount of data available that emphasizes the *intelligent* aspect of access—no one is willing to or capable of browsing through but a very small subset of the data collection, carefully selected to satisfy one's precise information need.

Research in artificial intelligence has long aimed at endowing machines with the ability to understand natural language. One of the core issues of this challenge is how to represent language semantics in a way that can be manipulated by computers. Prior work on semantics representation was based on purely statistical techniques, lexicographic knowledge, or elaborate endeavors to manually encode large amounts of knowledge. The simplest approach to represent the text semantics is to treat the text as an unordered bag of words, where the words themselves (possibly stemmed) become features of the textual object. The sheer ease of this approach makes it a reasonable candidate for many information retrieval tasks such as search and text categorization (Baeza-Yates & Ribeiro-Neto, 1999; Sebastiani, 2002). However, this simple model can only be reasonably used when texts are fairly long, and performs sub-optimally on short texts. Furthermore, it does little to address the two main problems of natural language processing (NLP), polysemy and synonymy.

Latent Semantic Analysis (LSA) (Deerwester, Dumais, Furnas, Landauer, & Harshman, 1990) is another purely statistical technique, which leverages word co-occurrence information from a large unlabeled corpus of text. LSA does not use any explicit human-organized knowledge; rather, it "learns" its representation by applying Singular Value Decomposition (SVD) to the words-by-documents co-occurrence matrix. LSA is essentially a dimensionality reduction technique that identifies a number of most prominent dimensions in the data, which are assumed to correspond to "latent concepts". Meanings of words and documents are then represented in the space defined by these concepts.

Lexical databases such as WordNet (Fellbaum, 1998) or Roget's Thesaurus (Roget, 1852) encode important relations between words such as synonymy, hypernymy, and meronymy. Approaches based on such resources (Budanitsky & Hirst, 2006; Jarmasz, 2003) map text words into word senses, and use the latter as concepts. However, lexical resources offer little information about the different word senses, thus making word sense disambiguation nearly impossible to achieve. Another drawback of such approaches is that creation of lexical resources requires lexicographic expertise as well as a lot of time and effort, and consequently such resources cover only a small fragment of the language lexicon. Specifically, such resources contain few proper names, neologisms, slang, and domain-specific technical terms. Furthermore, these resources have strong lexical orientation in that they predominantly contain information about individual words, but little world knowledge in general. Being inherently limited to individual words, these approaches require an extra level of sophistication to handle longer texts (Mihalcea, Corley, & Strapparava, 2006); for example, computing the similarity of a pair of texts amounts to comparing each word of one text to each word of the other text.

Studies in artificial intelligence have long recognized the importance of knowledge for problem solving in general, and for natural language processing in particular. Back in the early years of AI research, Buchanan and Feigenbaum (1982) formulated the *knowledge as power hypothesis*, which postulated that "The power of an intelligent program to perform its task well depends primarily on the quantity and quality of knowledge it has about that task."

When computer programs face tasks that require human-level intelligence, such as natural language processing, it is only natural to use an encyclopedia to endow the machine with the breadth of knowledge available to humans. There are, however, several obstacles on the way to using encyclopedic knowledge. First, such knowledge is available in textual form, and using it may require natural language understanding, a major problem in its own right. Furthermore, even language understanding may not be enough, as texts written *for* humans normally assume the reader possesses a large amount of common-sense knowledge, which is omitted even from most detailed encyclopedia articles (Lenat, 1997). Thus, there is a circular dependency—understanding encyclopedia articles requires natural language understanding capabilities, while the latter in turn require encyclopedic knowledge. To address this situation, Lenat and his colleagues launched the CYC project, which aims to explicitly catalog the common sense knowledge of the humankind.

We developed a new methodology that makes it possible to use an encyclopedia *directly*, without the need for manually encoded common-sense knowledge. Observe that an encyclopedia consists of a large collection of articles, each of which provides a comprehensive exposition focused on a single topic. Thus, we view an encyclopedia as a collection of *con-* *cepts* (corresponding to articles), each accompanied with a large body of text (the article contents). We propose to use the high-dimensional space defined by these concepts in order to represent the meaning of natural language texts. Compared to the bag of words and LSA approaches, using these concepts allows the computer to benefit from huge amounts of world knowledge, which is normally accessible to humans. Compared to electronic dictionaries and thesauri, our method uses knowledge resources that are over an order of magnitude larger, and also uniformly treats texts that are arbitrarily longer than a single word. Even more importantly, our method uses the body of text that accompanies the concepts in order to perform word sense disambiguation. As we show later, using the knowledge-rich concepts addresses both polysemy and synonymy, as we no longer manipulate mere words. We call our method Explicit Semantic Analysis (ESA), as it uses knowledge concepts explicitly defined and manipulated by humans.

Our approach is applicable to many NLP tasks whose input is a document (or shorter natural language utterance), and the output is a decision based on the document contents. Examples of such tasks are information retrieval (whether the document is relevant), text categorization (whether the document belongs to a certain category), or comparing pairs of documents to assess their similarity.[1]

Observe that documents manipulated in these tasks are given *in the same form* as the encyclopedic knowledge we intend to use—plain text. It is this key observation that allows us to circumvent the obstacles we enumerated above, and use the encyclopedia directly, without the need for deep language understanding or pre-cataloged common-sense knowledge. We quantify the degree of relevance of each Wikipedia concept to the input text by comparing this text to the article associated with the concept.

Let us illustrate the importance of external knowledge with a couple of examples. Without using external knowledge (specifically, knowledge about financial markets), one can infer little information from a very brief news title "Bernanke takes charge". However, using the algorithm we developed for consulting Wikipedia, we find that the following concepts are highly relevant to the input: BEN BERNANKE, FEDERAL RESERVE, CHAIRMAN OF THE FEDERAL RESERVE, ALAN GREENSPAN (Bernanke's predecessor), MONETARISM (an economic theory of money supply and central banking), INFLATION and DEFLATION. As another example, consider the title "Apple patents a Tablet Mac". Without deep knowledge of hi-tech industry and gadgets, one finds it hard to predict the contents of the news item. Using Wikipedia, we identify the following related concepts: APPLE COMPUTER[2], MAC OS (the Macintosh operating system) LAPTOP (the general name for portable computers, of which Tablet Mac is a specific example), AQUA (the GUI of MAC OS X), IPOD (another prominent product by Apple), and APPLE NEWTON (the name of Apple's early personal digital assistant).

For ease of presentation, in the above examples we only showed a few concepts identified by ESA as the most relevant for the input. However, the essence of our method is representing the meaning of text as a weighted combination of *all* Wikipedia concepts. Then, depending on the nature of the task at hand we either use these entire vectors of concepts, or use a few most relevant concepts to enrich the bag of words representation.

---

1. Thus, we do not consider tasks such as machine translation or natural language generation, whose output includes a new piece of text based on the input.
2. Note that we correctly identify the concept representing the computer company (APPLE COMPUTER) rather than the fruit (APPLE).

The contributions of this paper are twofold. First, we propose a new methodology to use Wikipedia for enriching representation of natural language texts. Our approach, named Explicit Semantic Analysis, effectively capitalizes on human knowledge encoded in Wikipedia, leveraging information that cannot be deduced solely from the input texts being processed. Second, we evaluate ESA on two commonly occurring NLP tasks, namely, text categorization and computing semantic relatedness of texts. In both tasks, using ESA resulted in significant improvements over the existing state of the art performance.

Recently, ESA was used by other researchers in a variety of tasks, and consistently proved to be superior to approaches that do not explicitly used large-scale repositories of human knowledge. Gurevych, Mueller, and Zesch (2007) re-implemented our ESA approach for the German-language Wikipedia, and found it to be superior for judging semantic relatedness of words compared to a system based on the German version of WordNet (GermaNet). Chang, Ratinov, Roth, and Srikumar (2008) used ESA for a text classification task without explicit training set, learning only from the knowledge encoded in Wikipedia. Milne and Witten (2008) found ESA to compare favorably to approaches that are solely based on hyperlinks, thus confirming that the wealth of textual descriptions in Wikipedia is exlicitly superior to using structural information alone.

## 2. Explicit Semantic Analysis

What is the meaning of the word "cat"? One way to interpret the word "cat" is via an explicit definition: a cat is a mammal with four legs, which belongs to the feline species, etc. Another way to interpret the meaning of "cat" is by the strength of its association with concepts that we know: "cat" relates strongly to the concepts "feline" and "pet", somewhat less strongly to the concepts "mouse" and "Tom & Jerry", etc.

We use this latter association-based method to assign semantic interpretation to words and text fragments. We assume the availability of a vector of *basic concepts*, $C_1, \ldots, C_n$, and we represent each text fragment $t$ by a vector of weights, $w_1, \ldots, w_n$, where $w_i$ represents the strength of association between $t$ and $C_i$. Thus, the set of basic concepts can be viewed as a canonical $n$-dimensional semantic space, and the semantics of each text segment corresponds to a point in this space. We call this weighted vector the *semantic interpretation vector* of $t$.

Such a canonical representation is very powerful, as it effectively allows us to estimate semantic relatedness of text fragments by their distance in this space. In the following section we describe the two main components of such a scheme: the set of *basic concepts*, and the algorithm that maps text fragments into interpretation vectors.

### 2.1 Using Wikipedia as a Repository of Basic Concepts

To build a general semantic interpreter that can represent text meaning for a variety of tasks, the set of basic concepts needs to satisfy the following requirements:

1. It should be comprehensive enough to include concepts in a large variety of topics.

2. It should be constantly maintained so that new concepts can be promptly added as needed.

3. Since the ultimate goal is to interpret *natural* language, we would like the concepts to be *natural*, that is, concepts recognized and used by human beings.

4. Each concept $C_i$ should have associated text $d_i$, so that we can determine the strength of its affinity with each term in the language.

Creating and maintaining such a set of natural concepts requires enormous effort of many people. Luckily, such a collection already exists in the form of Wikipedia, which is one of the largest knowledge repositories on the Web. Wikipedia is available in dozens of languages, while its English version is the largest of all, and contains 300+ million words in nearly one million articles, contributed by over 160,000 volunteer editors. Even though Wikipedia editors are not required to be established researchers or practitioners, the open editing approach yields remarkable quality. A recent study (Giles, 2005) found Wikipedia accuracy to rival that of Encyclopaedia Britannica. However, Britannica is about an order of magnitude smaller, with 44 million words in 65,000 articles (`http://store.britannica.com`, visited on February 10, 2006).

As appropriate for an encyclopedia, each article comprises a comprehensive exposition of a single topic. Consequently, we view each Wikipedia article as defining a concept that corresponds to each topic. For example, the article about artificial intelligence defines the concept Artificial Intelligence, while the article about parasitic extraction in circuit design defines the concept Layout extraction.[3] The body of the articles is critical in our approach, as it allows us to compute the affinity between the concepts and the words of the input texts.

An important advantage of our approach is thus the use of vast amounts of highly organized human knowledge. Compared to lexical resources such as WordNet, our methodology leverages knowledge bases that are orders of magnitude larger and more comprehensive. Importantly, the Web-based knowledge repositories we use in this work undergo constant development so their breadth and depth steadily increase over time. Compared to Latent Semantic Analysis, our methodology explicitly uses the knowledge collected and organized by humans. Our semantic analysis is *explicit* in the sense that we manipulate manifest concepts grounded in human cognition, rather than "latent concepts" used by LSA. Therefore, we call our approach Explicit Semantic Analysis (ESA).

## 2.2 Building a Semantic Interpreter

Given a set of concepts, $C_1, \ldots, C_n$, and a set of associated documents, $d_1, \ldots, d_n$, we build a sparse table $T$ where each of the $n$ columns corresponds to a concept, and each of the rows corresponds to a word that occurs in $\bigcup_{i=1\ldots n} d_i$. An entry $T[i, j]$ in the table corresponds to the TFIDF value of term $t_i$ in document $d_j$

$$T[i, j] = tf(t_i, d_j) \cdot \log \frac{n}{df_i},$$

---

3. Here we use the titles of articles as a convenient way to refer to the articles, but our algorithm treats the articles as atomic concepts.

where *term frequency* is defined as

$$tf(t_i, d_j) = \begin{cases} 1 + \log count(t_i, d_j), & if\ count(t_i, d_j) > 0 \\ 0, & otherwise \end{cases},$$

and $df_i = |\{d_k : t_i \in d_k\}|$ is the number of documents in the collection that contain the term $t_i$ (*document frequency*).

Finally, cosine normalization is applied to each row to disregard differences in document length:

$$T[i,j] \leftarrow \frac{T[i,j]}{\sqrt{\sum_{l=1}^{r} T[i,j]^2}},$$

where $r$ is the number of terms.

The semantic interpretation of a word $t_i$ is obtained as row $i$ of table $T$. That is, the meaning of a word is given by a vector of concepts paired with their TFIDF scores, which reflect the *relevance* of each concept to the word.

The semantic interpretation of a text fragment, $\langle t_1, \ldots, t_k \rangle$, is the centroid of the vectors representing the individual words. This definition allows us to partially perform word sense disambiguation. Consider, for example, the interpretation vector for the term "mouse". It has two sets of strong components, which correspond to two possible meanings: "mouse (rodent)" and "mouse (computing)". Similarly, the interpretation vector of the word "screen" has strong components associated with "window screen" and "computer screen". In a text fragment such as "I purchased a mouse and a screen", summing the two interpretation vectors will boost the computer-related components, effectively disambiguating both words.

Table T can also be viewed as an *inverted index*, which maps each word into a list of concepts in which it appears. Inverted index provides for very efficient computation of distance between interpretation vectors.

Given the amount of information encoded in Wikipedia, it is essential to control the amount of noise present in its text. We do so by discarding insufficiently developed articles, and by eliminating spurious association between articles and words. This is done by setting to zero the weights of those concepts whose weights for a given term are too low (see Section 3.2.3).

## 2.3 Using the Link Structure

It is only natural for an electronic encyclopedia to provide cross-references in the form of hyperlinks. As a result, a typical Wikipedia article has many more links to other entries than articles in conventional printed encyclopedias.

This link structure can be used in a number of ways. Observe that each link is associated with an *anchor text* (clickable highlighted phrase). The anchor text is not always identical to the canonical name of the target article, and different anchor texts are used to refer to the same article in different contexts. For example, anchor texts pointing at FEDERAL RESERVE include "Fed", "U.S. Federal Reserve Board", "U.S. Federal Reserve System", "Board of Governors of the Federal Reserve", "Federal Reserve Bank", "foreign reserves" and "Free Banking Era". Thus, anchor texts provide alternative names, variant spellings, and related phrases for the target concept, which we use to enrich the article text for the target concept.

Furthermore, inter-article links often reflect important relations between concepts that correspond to the linked articles. We explore the use of such relations for feature generation in the next section.

### 2.3.1 SECOND-ORDER INTERPRETATION

Knowledge concepts can be subject to many relations, including generalization, meronymy ("part of"), holonymy and synonymy, as well as more specific relations such as "capital of", "birthplace/birthdate of" etc. Wikipedia is a notable example of a knowledge repository that features such relations, which are represented by the hypertext links between Wikipedia articles.

These links encode a large amount of knowledge, which is not found in article texts. Consequently, leveraging this knowledge is likely to lead to better interpretation models. We therefore distinguish between *first-order* models, which only use the knowledge encoded in Wikipedia articles, and *second-order* models, which also incorporate the knowledge encoded in inter-article links. Similarly, we refer to the information obtained through inter-article links as second-order information.

As a rule, the presence of a link implies some relation between the concepts it connects. For example, the article on the UNITED STATES links to WASHINGTON, D.C. (country capital) and NORTH AMERICA (the continent where the country is situated). It also links to a multitude of other concepts, which are definitely related to the source concept, albeit it is more difficult to define those relations; these links include UNITED STATES DECLARATION OF INDEPENDENCE, PRESIDENT OF THE UNITED STATES, and ELVIS PRESLEY.

However, our observations reveal that the existence of a link does not always imply the two articles are strongly related.[4] In fact, many words and phrases in a typical Wikipedia article link to other articles just because there are entries for the corresponding concepts. For example, the Education subsection in the article on the UNITED STATES has gratuitous links to concepts HIGH SCHOOL, COLLEGE, and LITERACY RATE. Therefore, in order to use Wikipedia links for semantic interpretation, it is essential to filter the linked concepts according to their relevance to the text fragment being interpreted.

An intuitive way to incorporate concept relations is to examine a number of top-scoring concepts, and to boost the scores of the concepts linked from them. Let $ESA^{(1)}(t) = \left\langle w_1^{(1)}, \ldots, w_n^{(1)} \right\rangle$ be the interpretation vector of term $t$. We define the *second-level* interpretation of term $t$ as

$$ESA^{(2)}(t) = \left\langle w_1^{(2)}, \ldots, w_n^{(2)} \right\rangle$$

where

$$w_i^{(2)} = w_i^{(1)} + \alpha \cdot \sum_{\{j \mid \exists \text{link}(c_j, c_i)\}} w_j^{(1)}$$

Using $\alpha < 1$ ensures that the linked concepts are taken with reduced weights. In our experiments we used $\alpha = 0.5$.

---

4. The opposite is also true—the absence of a link may simply be due to an oversight. Adafre and de Rijke (2005) studied the problem of discovering missing links in Wikipedia.

### 2.3.2 Concept Generality Filter

Not all the new concepts identified through links are equally useful. Relevance of the newly added concepts is certainly important, but is not the only criterion. Suppose that we are given an input text "Google search". Which additional concept is likely to be more useful to characterize the input: Nigritude ultramarine (a specially crafted meaningless phrase used in a search engine optimization contest) or Website? Now suppose the input is "artificial intelligence" — which concept is likely to contribute more to the representation of this input, John McCarthy (computer scientist) or Logic? We believe that in both examples, the second concept would be more useful because it is not overly specific.

Consequently, we conjecture that we should add linked concepts sparingly, taking only those that are "more general" than the concepts that triggered them. But how can we judge the generality of concepts? While this may be tricky to achieve in the general case (no pun intended), we propose the following task-oriented criterion. Given two concepts $c_a$ and $c_b$, we compare the numbers of links pointing at them. Then, we say that $c_a$ is "more general" than $c_b$ if its number of incoming links is at least an order of magnitude larger, that is, if $log_{10}(\#inlinks(c_a)) - log_{10}(\#inlinks(c_b)) > 1$.

We show examples of additional concepts identified using inter-article links in Section 4.5.1. In Section 4.5.4 we evaluate the effect of using inter-article links as an additional knowledge source. In this section we also specifically examine the effect of only using more general linked concepts (i.e., adding concepts that are more general than the concepts that triggered them).

## 3. Using Explicit Semantic Analysis for Computing Semantic Relatedness of Texts

In this section we discuss the application of our semantic interpretation methodology to automatic assessment of semantic relatedness of words and texts.[5]

### 3.1 Automatic Computation of Semantic Relatedness

How related are "cat" and "mouse"? And what about "preparing a manuscript" and "writing an article"? The ability to quantify semantic relatedness of texts underlies many fundamental tasks in computational linguistics, including word sense disambiguation, information retrieval, word and text clustering, and error correction (Budanitsky & Hirst, 2006). Reasoning about semantic relatedness of natural language utterances is routinely performed by humans but remains an unsurmountable obstacle for computers. Humans do not judge text relatedness merely at the level of text words. Words trigger reasoning at a much deeper level that manipulates *concepts*—the basic units of meaning that serve humans to organize and share their knowledge. Thus, humans interpret the specific wording of a document in the much larger context of their background knowledge and experience. Lacking such elaborate resources, computers need alternative ways to represent texts and reason about them.

Explicit Semantic Analysis represents text as interpretation vectors in the high-dimensional space of concepts. With this representation, computing semantic relatedness of texts

---

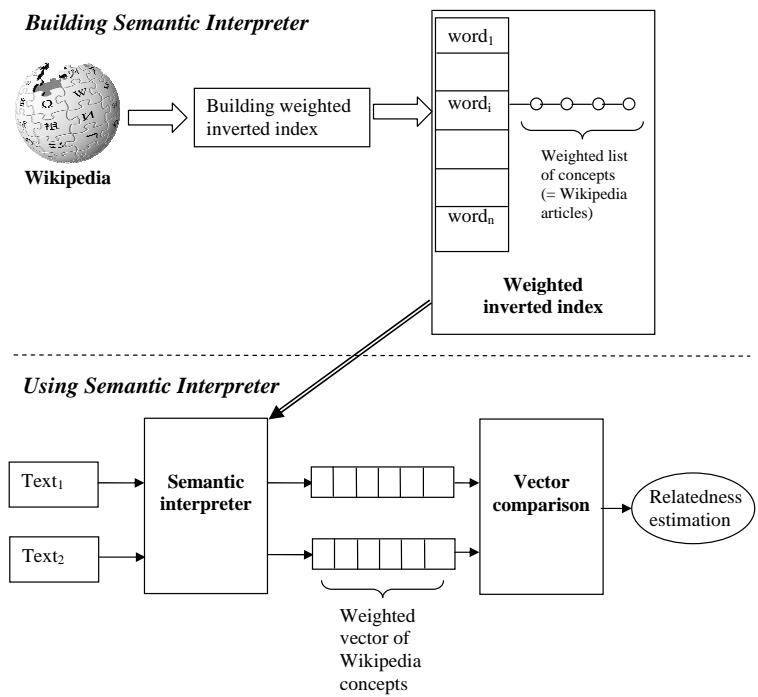

Figure 1: Knowledge-based semantic interpreter

simply amounts to comparing their vectors. Vectors could be compared using a variety of metrics (Zobel & Moffat, 1998); we use the cosine metric throughout the experiments reported in this paper. Figure 1 illustrates this process.

## 3.2 Implementation Details

We used Wikipedia snapshot as of November 11, 2005. After parsing the Wikipedia XML dump, we obtained 1.8 Gb of text in 910,989 articles. Although Wikipedia has almost a million articles, not all of them are equally useful for feature generation. Some articles correspond to overly specific concepts (e.g., METNAL, the ninth level of the Mayan underworld), or are otherwise unlikely to be useful for subsequent text categorization (e.g., specific dates or a list of events that happened in a particular year). Other articles are just too short, so we cannot reliably classify texts onto the corresponding concepts. We developed a set of simple heuristics for pruning the set of concepts, by discarding articles that have fewer than 100 non stop words or fewer than 5 incoming and outgoing links. We also discard articles that describe specific dates, as well as Wikipedia disambiguation pages, category pages and the like. After the pruning, 171,332 articles were left that defined concepts used for feature generation. We processed the text of these articles by first tokenizing it, removing stop words and rare words (occurring in fewer than 3 articles), and stemmed the remaining words; this yielded 296,157 distinct terms.

### 3.2.1 Preprocessing of Wikipedia XML Dump

Wikipedia data is publicly available online at `http://download.wikimedia.org`. All the data is distributed in XML format, and several packaged versions are available: article texts, edit history, list of page titles, interlanguage links etc. In this project, we only use the article texts, but ignore the information on article authors and page modification history. Before building the semantic interpreter, we perform a number of operations on the distributed XML dump:

- We simplify the original XML by removing all those fields that are not used in feature generation, such as author ids and last modification times.

- Wikipedia syntax defines a proprietary format for inter-article links, whereas the name of the article referred to is enclosed in brackets (e.g., "[UNITED STATES]"). We map all articles to numeric ids, and for each article build a list of ids of the articles it refers to. We also count the number of incoming and outgoing links for each article.

- Wikipedia defines a *redirection* mechanism, which maps frequently used variant names of entities into canonical names. For examples, UNITED STATES OF AMERICA is mapped to UNITED STATES. We resolve all such redirections during initial preprocessing.

- Another frequently used mechanism is *templates*, which allows articles to include frequently reused fragments of text without duplication, by including pre-defined and optionally parameterized templates on the fly. To speed up subsequent processing, we resolve all template inclusions at the beginning.

- We also collect all anchor texts that point at each article.

This preprocessing stage yields a new XML file, which is then used for building the feature generator.

### 3.2.2 The Effect of Knowledge Breadth

Wikipedia is being constantly expanded with new material as volunteer editors contribute new articles and extend the existing ones. Consequently, we conjectured that such addition of information should be beneficial for ESA, as it would rely on a larger knowledge base.

To test this assumption, we also acquired a newer Wikipedia snapshot as of March 26, 2006. Table 1 presents a comparison in the amount of information between two Wikipedia snapshots we used. The number of articles shown in the table reflects the total number of articles as of the date of the snapshot. The next table line (the number of concepts used) reflects the number of concepts that remained after the pruning as explained in the beginning of Section 3.2.

In the following sections we will confirm that using a larger knowledge base is beneficial for ESA, by juxtaposing the results obtained with the two Wikipedia snapshots. Therefore, no further dimensionality reduction is performed, and each input text fragment is represented in the space of up to 171,332 features (or 241,393 features in the case of the later Wikipedia snapshot); of course, many of the features will have zero values, so the feature vectors are sparse.

| | Wikipedia snapshot as of November 11, 2005 | Wikipedia snapshot as of March 23, 2006 |
|---|---|---|
| Combined article text | 1.8 Gb | 2.9 Gb |
| Number of articles | 910,989 | 1,187,839 |
| Concepts used | 171,332 | 241,393 |
| Distinct terms | 296,157 | 389,202 |

Table 1: Comparison of two Wikipedia snapshots

### 3.2.3 INVERTED INDEX PRUNING

We eliminate spurious association between articles and words by setting to zero the weights of those concepts whose weights for a given term are too low.

The algorithm for pruning the inverted index operates as follows. We first sort all the concepts for a given word according to their TFIDF weights in decreasing order. We then scan the resulting sequence of concepts with a sliding window of length 100, and truncate the sequence when the difference in scores between the first and last concepts in the window drops below 5% of the highest-scoring concept for this word (which is positioned first in the sequence). This technique looks for fast drops in the concept scores, which would signify that the concepts in the tail of the sequence are only loosely associated with the word (i.e., even though the word occurred in the articles corresponding to these concepts, it its not truly characteristic of the article contents). We evaluated more principled approaches such observing the values of the first and second derivatives, but the data seemed to be too noisy for reliable estimation of derivatives. Other researchers studied the use of derivatives in similar contexts (e.g., Begelman, Keller, & Smadja, 2006), and also found that the derivative alone is not sufficient, hence they found it necessary to estimate the magnitude of peaks by other means. Consequently, we opted to use the simple and efficient metric.

The purpose of such pruning is to eliminate spurious associations between concepts and terms, and is mainly beneficial for pruning the inverted index entries for very common words that occur in many Wikipedia articles. Using the above criteria, we analyzed the inverted index for the Wikipedia version dated November 11, 2005 (see Section 3.2.2). For the majority of terms, there were either fewer than 100 concepts with non-zero weight, or the concept-term weights decreased gracefully and did not qualify for pruning. We pruned the entries of 4866 terms out of the total of 296,157 terms. Among the terms whose concept vector was pruned, the term "link" had the largest number of concepts with non-zero weight—106,988—of which we retained only 838 concepts (0.8%); as another example, the concept vector for the term "number" was pruned from 52,244 entries down to 1360 (2.5%). On the average, 24% of concepts have been retained. The pruning rates for the second Wikipedia version (dated March 23, 2006) have been similar to these.

### 3.2.4 PROCESSING TIME

Using world knowledge requires additional computation. This extra computation includes the (one-time) preprocessing step where the semantic interpreter is built, as well as the actual mapping of input texts into interpretation vectors, performed online. On a standard workstation, parsing the Wikipedia XML dump takes about 7 hours, and building the semantic interpreter takes less than an hour. After the semantic interpreter is built, its throughput (i.e., the generation of interpretation vectors for textual input) is several hundred words per second. In the light of the improvements computing semantic relatedness and in text categorization accuracy that we report in Sections 3 and 4, we believe that the extra processing time is well compensated for.

## 3.3 Empirical Evaluation of Explicit Semantic Analysis

Humans have an innate ability to judge semantic relatedness of texts. Human judgements on a reference set of text pairs can thus be considered correct by definition, a kind of "gold standard" against which computer algorithms are evaluated. Several studies measured inter-judge correlations and found them to be consistently high (Budanitsky & Hirst, 2006; Jarmasz, 2003; Finkelstein, Gabrilovich, Matias, Rivlin, Solan, Wolfman, & Ruppin, 2002a), $r = 0.88 - 0.95$. These findings are to be expected—after all, it is this consensus that allows people to understand each other. Consequently, our evaluation amounts to computing the correlation of ESA relatedness scores with human judgments.

To better evaluate Wikipedia-based semantic interpretation, we also implemented a semantic interpreter based on another large-scale knowledge repository—the Open Directory Project (ODP, `http://www.dmoz.org`), which is the largest Web directory to date. In the case of ODP, concepts $C_i$ correspond to categories of the directory (e.g., Top/Computers/Artificial Intelligence), and text $d_i$ associated with each concept is obtained by pooling together the titles and descriptions of the URLs catalogued under the corresponding category. Interpretation of a text fragment amounts to computing a weighted vector of ODP concepts, ordered by their affinity to the input text. We built the ODP-based semantic interpreter using an ODP snapshot as of April 2004. Further implementation details can be found in our previous work (Gabrilovich & Markovitch, 2005, 2007b).

### 3.3.1 Test Collections

In this work, we use two datasets that to the best of our knowledge are the largest publicly available collections of their kind.[6] For both test collections, we use the correlation of computer-assigned scores with human scores to assess the algorithm performance.

To assess word relatedness, we use the WordSimilarity-353 collection (Finkelstein et al., 2002a; Finkelstein, Gabrilovich, Matias, Rivlin, Solan, Wolfman, & Ruppin, 2002b), which contains 353 noun pairs representing various degrees of similarity.[7] Each pair has 13–16 human judgements made by individuals with university degrees having either mother-tongue-level or otherwise very fluent command of the English language. Word pairs were assigned relatedness scores on the scale from 0 (totally unrelated words) to 10 (very much related or identical words). Judgements collected for each word pair were then averaged to produce a single relatedness score.[8] Spearman's rank-order correlation coefficient was used to compare computed relatedness scores with human judgements; being non-parametric, Spearman's correlation coefficient is considered to be much more robust than Pearson's linear correlation. When comparing our results to those of other studies, we have computed the Spearman's correlation coefficient with human judgments based on their raw data.

---

6. Recently, Zesch and Gurevych (2006) discussed automatic creation of datasets for assessing semantic similarity. However, the focus of their work was on automatical generation of a set of sufficiently diverse word pairs, thus relieving the humans of the need to construct word lists manually. Obviously, establishing the "gold standard" semantic relatedness for each word pair is still performed manually by human judges.

7. Some previous studies (Jarmasz & Szpakowicz, 2003) suggested that the word pairs comprising this collection might be culturally biased.

For document similarity, we used a collection of 50 documents from the Australian Broadcasting Corporation's news mail service (Lee, Pincombe, & Welsh, 2005; Pincombe, 2004). The documents were between 51 and 126 words long, and covered a variety of topics. The judges were 83 students from the University of Adelaide, Australia, who were paid a small fee for their work. These documents were paired in all possible ways, and each of the 1,225 pairs has 8–12 human judgements (averaged for each pair). To neutralize the effects of ordering, document pairs were presented in random order, and the order of documents within each pair was randomized as well. When human judgements have been averaged for each pair, the collection of 1,225 relatedness scores have only 67 distinct values. Spearman's correlation is not appropriate in this case, and therefore we used Pearson's linear correlation coefficient.

Importantly, instructions for human judges in both test collections specifically directed the participants to assess the *degree of relatedness* of words and texts involved. For example, in the case of antonyms, judges were instructed to consider them as "similar" rather than "dissimilar".

### 3.3.2 Prior Work

A number of prior studies proposed a variety of approaches to computing word similarity using WordNet, Roget's thesaurus, and LSA. Table 2 presents the results of applying these approaches to the WordSimilarity-353 test collection.

Jarmasz (2003) replicated the results of several WordNet-based methods, and compared them to a new approach based on Roget's Thesaurus. Hirst and St-Onge (1998) viewed WordNet as a graph, and considered the length and directionality of the graph path connecting two nodes. Leacock and Chodorow (1998) also used the length of the shortest graph path, and normalized it by the maximum taxonomy depth. Jiang and Conrath (1997), and later Resnik (1999), used the notion of information content of the lowest node subsuming two given words. Lin (1998b) proposed a computation of word similarity based on the information theory. See (Budanitsky & Hirst, 2006) for a comprehensive discussion of WordNet-based approaches to computing word similarity.

According to Jarmasz (2003), Roget's Thesaurus has a number of advantages compared to WordNet, including links between different parts of speech, topical groupings, and a variety of relations between word senses. Consequently, the method developed by the authors using Roget's as a source of knowledge achieved much better results than WordNet-based methods. Finkelstein et al. (2002a) reported the results of computing word similarity using an LSA-based model (Deerwester et al., 1990) trained on the Grolier Academic American Encyclopedia. Recently, Hughes and Ramage (2007) proposed a method for computing semantic relatedness using random graph walks; their results on the WordSimilarity-353 dataset are competitive with those reported by Jarmasz (2003) and Finkelstein et al. (2002a).

---

8. Finkelstein et al. (2002a) report inter-judge agreement of 0.95 for the WordSimilarity-353 collection. We have also performed our own assessment of the inter-judge agreement for this dataset. Following Snow, O'Connor, Jurafsky, and Ng (2008), we divided the human judges into two sets and averaged the numeric judgements for each word pair among the judges in the set, thus yielding a (353 element long) vector of average judgments for each set. Spearman's correlation coefficient between the vectors of the two sets was 0.903.

Strube and Ponzetto (2006) proposed an alternative approach to computing word similarity based on Wikipedia, by comparing articles in whose titles the words occur. We discuss this approach in greater detail in Section 5.1.

Prior work on assessing the similarity of textual documents was based on comparing the documents as bags of words, as well as on LSA. Lee et al. (2005) compared a number of approaches based on the bag of words representation, which used both binary and *tfidf* representation of word weights and a variety of similarity measures (correlation, Jaccard, cosine, and overlap). The authors also implemented an LSA-based model trained on a set of news documents from the Australian Broadcasting Corporation (test documents whose similarity was computed came from the same distribution). The results of these experiments are reported in Table 3.

### 3.3.3 RESULTS

To better understand how Explicit Semantic Analysis works, let us consider similarity computation for pairs of actual phrases. For example, given two phrases "scientific article" and "journal publication", ESA determines that the following Wikipedia concepts are found among the top 20 concepts for each phrase: SCIENTIFIC JOURNAL, NATURE (JOURNAL), ACADEMIC PUBLICATION, SCIENCE (JOURNAL), and PEER REVIEW. When we compute similarity of "RNA" and "DNA", the following concepts are found to be shared among the top 20 lists: TRANSCRIPTION (GENETICS), GENE, RNA, and CELL (BIOLOGY). It is the presence of identical concepts among the top concepts characterizing each phrase that allows ESA to establish their semantic similarity.

Table 2 shows the results of applying our methodology to estimating relatedness of individual words, with statistically significant improvements shown in bold. The values shown in the table represent Spearman's correlation between the human judgments and the relatedness scores produced by the different methods. Jarmasz (2003) compared the performance of 5 WordNet-based metrics, namely, those proposed by Hirst and St-Onge (1998), Jiang and Conrath (1997), Leacock and Chodorow (1998), Lin (1998b), and Resnik (1999). In Table 2 we report the performance of the best of these metrics, namely, those by Lin (1998b) and Resnik (1999). In the WikiRelate! paper (Strube & Ponzetto, 2006), the authors report results of as many as 6 different method variations, and again we report the performance of the best one (based on the metric proposed by Leacock and Chodorow, 1998).

As we can see, both ESA techniques yield substantial improvements over previous state of the art results. Notably, ESA also achieves much better results than another recently introduce method based on Wikipedia (Strube & Ponzetto, 2006). We provide a detailed comparison of our approach with this latter work in Section 5.1. Table 3 shows the results for computing relatedness of entire documents. In both tables, we show the statistical significance of the difference between the performance of ESA-Wikipedia (March 26, 2006 version) and that of other algorithms[9] by using Fisher's $z$-transformation (Press, Teukolsky, Vetterling, & Flannery, 1997, Section 14.5).

| Algorithm | Spearman's correlation with human judgements | Stat. significance ($p$-value) |
|---|---|---|
| WordNet-based techniques (Jarmasz, 2003) | 0.35 | $\mathbf{4 \cdot 10^{-16}}$ |
| Roget's Thesaurus-based technique (Jarmasz, 2003) | 0.55 | $\mathbf{1.3 \cdot 10^{-6}}$ |
| LSA (Finkelstein et al., 2002a) | 0.56 | $\mathbf{3.4 \cdot 10^{-6}}$ |
| WikiRelate! (Strube & Ponzetto, 2006) | 0.50 | $\mathbf{8 \cdot 10^{-9}}$ |
| MarkovLink (Hughes & Ramage, 2007) | 0.55 | $\mathbf{1.6 \cdot 10^{-6}}$ |
| ESA-Wikipedia (March 26, 2006 version) | 0.75 | – |
| ESA-Wikipedia (November 11, 2005 version) | 0.74 | – |
| ESA-ODP | 0.65 | **0.0044** |

Table 2: Spearman's rank correlation of word relatedness scores with human judgements on the WordSimilarity-353 collection

| Algorithm | Pearson's correlation with human judgements | Stat. significance ($p$-value) |
|---|---|---|
| Bag of words (Lee et al., 2005) | 0.1–0.5 | $\mathbf{4 \cdot 10^{-19}}$ |
| LSA (Lee et al., 2005) | 0.60 | $\mathbf{5 \cdot 10^{-8}}$ |
| ESA-Wikipedia (March 26, 2006 version) | 0.72 | – |
| ESA-Wikipedia (November 11, 2005 version) | 0.71 | – |
| ESA-ODP | 0.69 | 0.07 |

Table 3: Pearson's correlation of text relatedness scores with human judgements on Lee et al.'s document collection

On both test collections, Wikipedia-based semantic interpretation is superior to the ODP-based one; in the word relatedness task, this superiority is statistically significant at $p < 0.005$. We believe that two factors contribute to this phenomenon. First, axes of a multi-dimensional interpretation space should ideally be as independent as possible. The hierarchical organization of the Open Directory reflects the generalization relation between concepts and obviously violates this independence requirement. Second, to increase the amount of training data for building the ODP-based semantic interpreter, we crawled all the URLs listed in the ODP. This allowed us to increase the amount of textual data by several orders of magnitude, but also brought about a non-negligible amount of noise, which is common in Web pages. On the other hand, Wikipedia articles are virtually noise-free, and mostly qualify as Standard Written English. Thus, the textual descriptions of Wikipedia concepts are arguably more focused than those of the ODP concepts.

---

9. Whenever a range of values is available, we compared ESA-Wikipedia with the best-performing method in the range.

It is also essential to note that in both experiments, using a newer Wikipedia snapshot leads to better results (although the difference between the performance of two versions is admittedly small).

We evaluated the effect of using second-order interpretation for computing semantic relatedness of texts, but it only yielded negligible improvements. We hypothesize that the reason for this finding is that computing semantic relatedness essentially uses all available Wikipedia concepts, so second-order interpretation can only slightly modify the weights of existing concepts. In the next section, which describes the application of ESA to text categorization, we trim the interpretation vectors for the sake of efficiency, and only consider a few highest-scoring concepts for each input text fragment. In this scenario, second-order interpretation does have a positive effect and actually improves the accuracy of text categorization (Section 4.5.4). This happens because only a few selected Wikipedia concepts are used to augment text representation, and the second-order approach selectively adds highly related concepts identified by analyzing Wikipedia links.

## 4. Using Explicit Semantic Analysis for Text Categorization

In this section we evaluate the benefits of using external knowledge for text categorization.[10]

### 4.1 Background on Text Categorization

*Text categorization (TC)* deals with assigning category labels to natural language documents. Categories come from a fixed set of labels (possibly organized in a hierarchy) and each document may be assigned one or more categories. Text categorization systems are useful in a wide variety of tasks, such as routing news and e-mail to appropriate corporate desks, identifying junk email, or correctly handling intelligence reports.

The majority of existing text classification systems represent text as a *bag of words*, and use a variant of the vector space model with various weighting schemes (Salton & McGill, 1983). Thus, the features commonly used in text classification are weighted occurrence frequencies of individual words. State-of-the-art systems for text categorization use a variety of induction techniques, such as support vector machines, $k$-nearest neighbor algorithm, and neural networks. The bag of words (BOW) method is very effective in easy to medium difficulty categorization tasks where the category of a document can be identified by several easily distinguishable keywords. However, its performance becomes quite limited for more demanding tasks, such as those dealing with small categories or short documents.

There have been various attempts to extend the basic BOW approach. Several studies augmented the bag of words with n-grams (Caropreso, Matwin, & Sebastiani, 2001; Peng & Shuurmans, 2003; Mladenic, 1998; Raskutti, Ferra, & Kowalczyk, 2001) or statistical language models (Peng, Schuurmans, & Wang, 2004). Others used linguistically motivated features based on syntactic information, such as that available from part-of-speech tagging or shallow parsing (Sable, McKeown, & Church, 2002; Basili, Moschitti, & Pazienza, 2000). Additional studies researched the use of word clustering (Baker & McCallum, 1998; Bekker-man, 2003; Dhillon, Mallela, & Kumar, 2003), neural networks (Jo, 2000; Jo & Japkowicz, 2005; Jo, 2006), as well as dimensionality reduction techniques such as LSA (Deerwester et al., 1990; Hull, 1994; Zelikovitz & Hirsh, 2001; Cai & Hofmann, 2003). However, these attempts had mostly limited success.

---

10. Preliminary results of this research have been reported by Gabrilovich and Markovitch (2006).

We believe that the bag of words approach is inherently limited, as it can only use those pieces of information that are explicitly mentioned in the documents, and only if the same vocabulary is consistently used throughout. The BOW approach cannot generalize over words, and consequently words in the testing document that never appeared in the training set are necessarily ignored. Nor can synonymous words that appear infrequently in training documents be used to infer a more general principle that covers all the cases. Furthermore, considering the words as an unordered bag makes it difficult to correctly resolve the sense of polysemous words, as they are no longer processed in their native context. Most of these shortcomings stem from the fact that the bag of words method has no access to the wealth of world knowledge possessed by humans, and is therefore easily puzzled by facts and terms that cannot be easily deduced from the training set.

## 4.2 Using ESA for Feature Generation

We propose a solution that augments the bag of words with knowledge-based features. Given a document to be classified, we would like to use ESA to represent the document text in the space of Wikipedia concepts. However, text categorization is crucially different from computing semantic relatedness (cf. Section 3) in two important respects.

**First**, computing semantic relatedness is essentially a "one-off" task, that is, given a particular pair of text fragments, we need to quantify their relatedness with no prior examples for this specific task. In such cases, the very words of the text fragments are likely to be of marginal usefulness, especially when the two fragments are one word long. This happens because all the data available to us is limited to the two input fragments, which in most cases share few words, if at all.

On the other hand, in supervised text categorization, one is usually given a collection of labeled text documents, from which one can induce a text categorizer. Consequently, words that occur in the training examples can serve as valuable features—this is how the bag of words approach was born. As we have observed in an earlier work (Gabrilovich & Markovitch, 2005, 2007b), it is ill-advised to completely replace the bag of words with generated concepts, and instead it is advantageous to enrich the bag of words. Rather, we opt to *augment* the bag of words with carefully selected knowledge concepts, which become new *features* of the document. We refer to this process as *feature generation*, because we actually construct new document features beyond those in the bag of words.

**Second**, enriching document representation for text categorization with all possible Wikipedia concepts is extremely expensive computationally, because a machine learning classifier will be learned in the augmented feature space. Such a representation obviously takes a lot of storage space, and cannot be processed efficiently because of the multitude of the concepts involved (whose number can easily reach hundreds of thousands). Therefore, in the text categorization task, we prune the interpretation vectors to only retain a number of highest-scoring concepts for each input text fragment.

**Using the multi-resolution approach to feature generation**   We believe that considering the document as a single unit can often be misleading: its text might be too diverse to be readily mapped to the right set of concepts, while notions mentioned only briefly may be overlooked. Instead, we partition the document into a series of non-overlapping segments (called *contexts*), and then generate features at this finer level. Each context is mapped into a number of Wikipedia concepts in the knowledge base, and pooling these concepts together to describe the entire document results in *multi-faceted* classification. This way, the resulting set of concepts represents the various aspects or sub-topics covered by the document.

Potential candidates for such contexts are simple sequences of words, or more linguistically motivated chunks such as sentences or paragraphs. The optimal resolution for document segmentation can be determined automatically using a validation set. In our earlier work (Gabrilovich & Markovitch, 2005, 2007b), we proposed a more principled *multi-resolution* approach that simultaneously partitions the document at several levels of linguistic abstraction (windows of words, sentences, paragraphs, up to taking the entire document as one big chunk), and performs feature generation at each of these levels. We rely on the subsequent *feature selection* step to eliminate extraneous features, preserving only those that genuinely characterize the document.

It is essential to emphasize that using the multi-resolution approach only makes sense when interpretation vectors are pruned to only retain a number of highest-scoring concepts for each context. As explained above, this is exactly the case for text categorization. Without such pruning, producing interpretation vectors for each context and then summing them up would be equivalent to simply multiplying the weight of each concept by a constant factor. In order to explain why the situation is different in the presence of pruning, let us consider an example. Suppose we have a long document that only mentions a particular topic $T$ in its last paragraph. Since this topic is not central to the document, the $N$ top-scoring concepts in the document's interpretation vector $I$ are unlikely to cover this topic. Although $T$ is likely to be covered by other concepts in $I$, those concepts have lower weight in $I$ and are going to be pruned. However, if we produce interpretation vectors also for each paragraph of the document, and retain $N$ highest-scoring concepts of each, then the concepts generated for the last paragraph will cover $T$. Consequently, $T$ will have representation in the joined set of concepts generated for the document. In many text categorization tasks, documents are labeled with a particular topic even if they mention the topic briefly, hence generating features describing such topics is very important.

**Feature generation**   Feature generation is performed prior to text categorization. Each document is transformed into a series of local contexts, which are then represented as interpretation vectors using ESA. The top ten concepts of all the vectors are pooled together, and give rise to the generated features of the document, which are added to the bag of words. Since concepts in our approach correspond to Wikipedia articles, constructed features also correspond to the articles. Thus, a set of features generated for a document can be viewed as representing a set of Wikipedia articles that are most relevant to the document contents. The constructed features are used in conjunction with the original bag of words. The resulting set optionally undergoes feature selection, and the most discriminative features are retained for document representation.

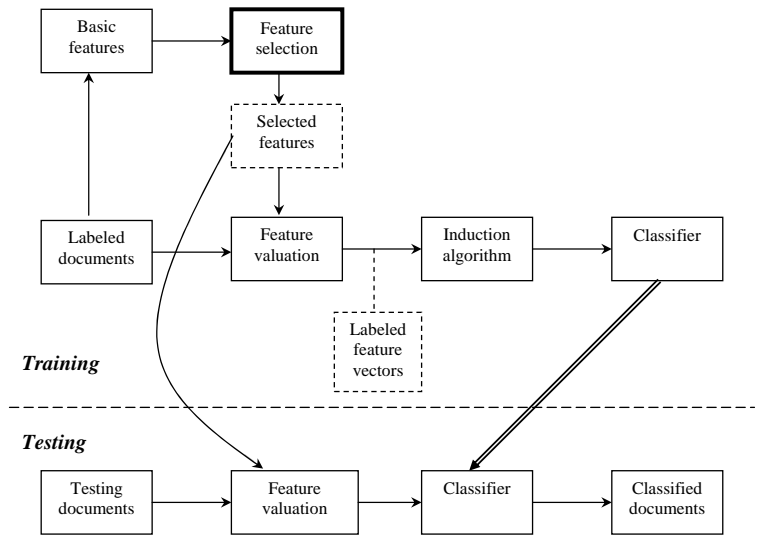

Figure 2: Standard approach to text categorization.

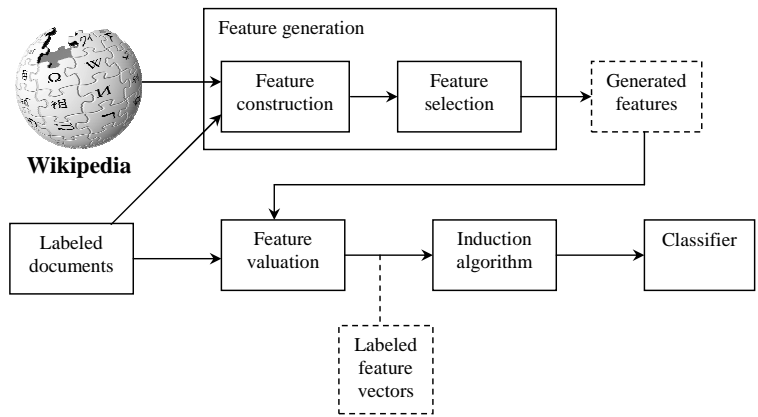

Figure 3: Induction of text classifiers using the proposed framework for feature generation.

Figure 2 depicts the standard approach to text categorization. Figure 3 outlines the proposed feature generation framework; observe that the "Feature generation" box replaces the "Feature selection" box framed in **bold** in Figure 2.

It is essential to note that we do *not* use the encyclopedia to simply increase the amount of the training data for text categorization; neither do we use it as a text corpus to collect word co-occurrence statistics. Rather, we use the knowledge distilled from the encyclopedia to enrich the representation of documents, so that a text categorizer is induced in the augmented, knowledge-rich feature space.

### 4.3 Test Collections

This section gives a brief description of the test collections we used to evaluate our methodology. We provide a much more detailed description of these test collections in Appendix B.

**1. Reuters-21578** (Reuters, 1997) is historically the most often used dataset in text categorization research. Following common practice, we used the ModApte split (9603 training, 3299 testing documents) and two category sets, 10 largest categories and 90 categories with at least one training and testing example.

**2. 20 Newsgroups (20NG)** (Lang, 1995) is a well-balanced dataset of 20 categories containing 1000 documents each.

**3. Movie Reviews (Movies)** (Pang, Lee, & Vaithyanathan, 2002) defines a sentiment classification task, where reviews express either positive or negative opinion about the movies. The dataset has 1400 documents in two categories (positive/negative).

**4. Reuters Corpus Volume I (RCV1)** (Lewis, Yang, Rose, & Li, 2004) has over 800,000 documents. To speed up the experiments, we used a subset of RCV1 with 17,808 training documents (dated 20–27/08/96) and 5,341 testing ones (28–31/08/96). Following Brank, Grobelnik, Milic-Frayling, and Mladenic (2002), we used 16 Topic and 16 Industry categories that constitute representative samples of the full groups of 103 and 354 categories, respectively. We also randomly sampled the Topic and Industry categories into 5 sets of 10 categories each.[11]

**5. OHSUMED** (Hersh, Buckley, Leone, & Hickam, 1994) is a subset of MEDLINE, which contains 348,566 medical documents. Each document contains a title, and about two-thirds (233,445) also contain an abstract. Each document is labeled with an average of 13 MeSH[12] categories (out of total 14,000). Following Joachims (1998), we used a subset of documents from 1991 that have abstracts, taking the first 10,000 documents for training and the next 10,000 for testing. To limit the number of categories for the experiments, we randomly generated 5 sets of 10 categories each.[13]

Using these 5 datasets allows us to comprehensively evaluate the performance of our approach. Specifically, comparing 20 Newsgroups and the two Reuters datasets (Reuters-21578 and Reuters Corpus Volume 1), we observe that the former is substantially more noisy since the data has been obtained from Usenet newsgroups, while the Reuters datasets are significantly cleaner. The *Movie Reviews* collection presents an example of *sentiment classification*, which is different from standard (topical) text categorization. Finally, the OHSUMED dataset presents an example of a very comprehensive taxonomy of over 14,000 categories. As we explain the next section, we also used this dataset to create a collection of labeled short texts, which allowed us to quantify the performance of our method on such texts.

**Short Documents** We also derived several datasets of short documents from the test collections described above. Recall that about one-third of OHSUMED documents have titles but no abstract, and can therefore be considered short documents "as-is." We used the same range of documents as defined above, but considered only those without abstracts; this yielded 4,714 training and 5,404 testing documents. For all other datasets, we created a short document from each original document by taking only the title of the latter (with the exception of Movie Reviews, where documents have no titles).

---

11. The full definition of the category sets we used is available in Table 8 (see Section B.4).
12. `http://www.nlm.nih.gov/mesh`
13. The full definition of the category sets we used is available in Table 9 (see Section B.5).

It should be noted, however, that substituting a title for the full document is a poor man's way to obtain a collection of classified short documents. When documents were first labeled with categories, the human labeller saw each document *in its entirety*. In particular, a category might have been assigned to a document on the basis of facts mentioned in its body, even though the information may well be missing from the (short) title. Thus, taking all the categories of the original documents to be "genuine" categories of the title is often misleading. However, because we know of no publicly available test collections of short documents, we decided to construct datasets as explained above. Importantly, OHSUMED documents without abstracts have been classified as such by humans; working with the OHSUMED-derived dataset can thus be considered a "pure" experiment.

## 4.4 Experimentation Procedure

We used support vector machines[14] as our learning algorithm to build text categorizers, since prior studies found SVMs to have the best performance for text categorization (Sebastiani, 2002; Dumais, Platt, Heckerman, & Sahami, 1998; Yang & Liu, 1999). Following established practice, we use the precision-recall break-even point (BEP) to measure text categorization performance. BEP is defined in terms of the standard measures of precision and recall, where precision is the proportion of true document-category assignments among all assignments predicted by the classifier, and recall is the proportion of true document-category assignments that were also predicted by the classifier. It is obtained by either tuning the classifier so that precision is equal to recall, or sampling several $(precision, recall)$ points that bracket the expected BEP value and then interpolating (or extrapolating, in the event that all the sampled points lie on the same side).

For the two Reuters datasets and OHSUMED we report both micro- and macro-averaged BEP, since their categories differ in size significantly. Micro-averaged BEP operates at the document level and is primarily affected by categorization performance on larger categories. On the other hand, macro-averaged BEP averages results for individual categories, and thus small categories with few training examples have large impact on the overall performance.

For both Reuters datasets (Reuters-21578 and RCV1) and OHSUMED we used a fixed train/test split as defined in Section 4.3, and consequently used macro sign test (S-test) (Yang & Liu, 1999) to assess the statistical significance of differences in classifier performance. For 20NG and Movies we performed 4-fold cross-validation, and used paired t-test to assess the significance. We also used the non-parametric Wilcoxon signed-ranks test (Demsar, 2006) to compare the baseline and the FG-based classifiers over multiple data sets. In the latter case, the individual measurements taken are the (micro- or macro-averaged) BEP values observed on each dataset.

---

14. We used the $SVM^{light}$ implementation (Joachims, 1999) with the default parameters. In our earlier work on feature selection (Gabrilovich & Markovitch, 2004), we conducted a thorough experimentation with a wide range of values of the C parameter, and found it not to be of any major importance for these datasets; consequently, we leave this parameter at its default setting as well.

### 4.4.1 Text Categorization Infrastructure

We conducted the experiments using a text categorization platform of our own design and development named $\mathcal{H}$OGWARTS[15] (Davidov, Gabrilovich, & Markovitch, 2004). We opted to build a comprehensive new infrastructure for text categorization, as surprisingly few software tools are publicly available for researchers, while those that are available allow only limited control over their operation. $\mathcal{H}$OGWARTS facilitates full-cycle text categorization including text preprocessing, feature extraction, construction, selection and weighting, followed by actual classification with cross-validation of experiments. The system currently provides XML parsing, part-of-speech tagging (Brill, 1995), sentence boundary detection, stemming (Porter, 1980), WordNet (Fellbaum, 1998) lookup, a variety of feature selection algorithms, and TFIDF feature weighting schemes. $\mathcal{H}$OGWARTS has over 250 configurable parameters that control its *modus operandi* in minute detail. $\mathcal{H}$OGWARTS interfaces with SVM, KNN and C4.5 text categorization algorithms, and computes all standard measures of categorization performance. $\mathcal{H}$OGWARTS was designed with a particular emphasis on processing efficiency, and portably implemented in the ANSI C++ programming language and C++ Standard Template Library. The system has built-in loaders for Reuters-21578 (Reuters, 1997), RCV1 (Lewis et al., 2004), 20 Newsgroups (Lang, 1995), Movie Reviews (Pang et al., 2002), and OHSUMED (Hersh et al., 1994), while additional datasets can be easily integrated in a modular way.

Each document undergoes the following processing steps. Document text is first tokenized, and title words are replicated twice to emphasize their importance. Then, stop words, numbers and mixed alphanumeric strings are removed, and the remaining words are stemmed. The bag of words is next merged with the set of features generated for the document by analyzing its contexts as explained in Section 4.2, and rare features occurring in fewer than 3 documents are removed.

Since earlier studies found that most BOW features are indeed useful for SVM text categorization[16] (Joachims, 1998; Rogati & Yang, 2002; Brank et al., 2002; Bekkerman, 2003; Leopold & Kindermann, 2002; Lewis et al., 2004), we take the bag of words in its entirety (with the exception of rare features removed in the previous step). The generated features, however, undergo feature selection using the information gain criterion.[17] Finally, feature weighting is performed using the "ltc" TF.IDF function (logarithmic term frequency and inverse document frequency, followed by cosine normalization) (Salton & Buckley, 1988; Debole & Sebastiani, 2003).

### 4.4.2 Baseline Performance of $\mathcal{H}$OGWARTS

We now demonstrate that the performance of basic text categorization in our implementation (column "Baseline" in Table 4) is consistent with the state of the art as reflected in other published studies (all using SVM). On Reuters-21578, Dumais et al. (1998) achieved micro-BEP of 0.920 for 10 categories and 0.870 for all categories. On 20NG[18], Bekkerman (2003) obtained BEP of 0.856. Pang et al. (2002) obtained accuracy of 0.829 on Movies[19]. The minor variations in performance are due to differences in data preprocessing in the different systems; for example, for the Movies dataset we worked with raw HTML files rather than with the official tokenized version, in order to recover sentence and paragraph structure for contextual analysis. For RCV1 and OHSUMED, direct comparison with published results is more difficult because we limited the category sets and the date span of documents to speed up experimentation.

---

15. *Hogwarts School of Witchcraft and Wizardry* is the educational institution attended by Harry Potter (Rowling, 1997).
16. Gabrilovich and Markovitch (2004) described a class of problems where feature selection from the bag of words actually improves SVM performance.
17. Of course, feature selection is performed using only the training set of documents.

### 4.4.3 Using the Feature Generator

The core engine of Explicit Semantic Analysis was implemented as explained in Section 3.2.

We used the *multi-resolution* approach to feature generation, classifying document contexts at the level of individual words, complete sentences, paragraphs, and finally the entire document.[20] For each context, features were generated from the 10 best-matching concepts produced by the feature generator.

## 4.5 Wikipedia-based Feature Generation

In this section, we report the results of an experimental evaluation of our methodology.

### 4.5.1 Qualitative Analysis of Feature Generation

We now study the process of feature generation on a number of actual examples.

**Feature Generation per se**  To illustrate our approach, we show features generated for several text fragments. Whenever applicable, we provide short explanations of the generated concepts; in most cases, the explanations are taken from Wikipedia (Wikipedia, 2006).

- **Text:** *"Wal-Mart supply chain goes real time"*

  **Top 10 generated features:** (1) Wal-Mart; (2) Sam Walton; (3) Sears Holdings Corporation; (4) Target Corporation; (5) Albertsons; (6) ASDA; (7) RFID; (8) Hypermarket; (9) United Food and Commercial Workers; (10) Chain store

  **Selected explanations:** (2) Wal-Mart founder; (5) prominent competitors of Wal-Mart; (6) a Wal-Mart subsidiary in the UK; (7) Radio Frequency Identification, a technology that Wal-Mart uses very extensively to manage its stock; (8) superstore (a general concept, of which Wal-Mart is a specific example); (9) a labor union that

---

18. For comparison with the results reported by Bekkerman (2003) we administered a single test run (i.e., without cross-validation), taking the first 3/4 of postings in each newsgroup for training, and the rest for testing.

19. For comparison with the results reported by Pang et al. (2002) we administered a single test run (i.e., without cross-validation), taking the first 2/3 of the data for each opinion type for training, and the rest for testing.

20. The 20NG dataset is an exception, owing to its high level of intrinsic noise that renders identification of sentence boundaries extremely unreliable, and causes word-level feature generation to produce too many spurious classifications. Consequently, for this dataset we restrict the multi-resolution approach to individual paragraphs and the entire document only.

has been trying to organize Wal-Mart's workers; (10) a general concept, of which Wal-Mart is a specific example

- It is particularly interesting to juxtapose the features generated for fragments that contain ambiguous words. To this end, we show features generated for two phrases that contain the word "bank" in two different senses, "Bank of America" (financial institution) and "Bank of Amazon" (river bank). As can be readily seen, our feature generation methodology is capable of performing word sense disambiguation by considering ambiguous words in the context of their neighbors.

  - **Text:** *"**Bank** of America"*

    **Top 10 generated features:** (1) BANK; (2) BANK OF AMERICA; (3) BANK OF AMERICA PLAZA (ATLANTA); (4) BANK OF AMERICA PLAZA (DALLAS); (5) MBNA (a bank holding company acquired by Bank of America); (6) VISA (CREDIT CARD); (7) BANK OF AMERICA TOWER, NEW YORK CITY; (8) NASDAQ; (9) MASTERCARD; (10) BANK OF AMERICA CORPORATE CENTER

  - **Text:** *"**Bank** of Amazon"*

    **Top 10 generated features:** (1) AMAZON RIVER; (2) AMAZON BASIN; (3) AMAZON RAINFOREST; (4) AMAZON.COM; (5) RAINFOREST; (6) ATLANTIC OCEAN; (7) BRAZIL; (8) LORETO REGION (a region in Peru, located in the Amazon Rainforest); (9) RIVER; (10) ECONOMY OF BRAZIL

- Our method, however, is not 100% accurate, and in some cases it generates features that are only somewhat relevant or even irrelevant to the input text. As an example, we show the outcome of feature generation for the title of our earlier article (Gabrilovich & Markovitch, 2006). For each concept, we show a list of input words that triggered it (the words are stemmed and sorted in the decreasing order of their contribution).

**Text:** *"Overcoming the Brittleness Bottleneck using Wikipedia: Enhancing Text Categorization with Encyclopedic Knowledge"*

**Top 10 generated features:**

1. ENCYCLOPEDIA (encyclopedia, knowledge, Wikipedia, text)
2. WIKIPEDIA (Wikipedia, enhance, encyclopedia, text)
3. ENTERPRISE CONTENT MANAGEMENT (category, knowledge, text, overcome, enhance)
4. PERFORMANCE PROBLEM (bottleneck, category, enhance)
5. IMMANUEL KANT (category, knowledge, overcome)
6. TOOTH ENAMEL (brittleness, text, enhance)
7. LUCID DREAMING (enhance, text, knowledge, category)
8. BOTTLENECK (bottleneck)
9. JAVA PROGRAMMING LANGUAGE (category, bottleneck, enhance)

10. Transmission Control Protocol (category, enhance, overcome)

Some of the generated features are clearly relevant to the input, such as Encyclopedia, Wikipedia, and Enterprise content management. Others, however, are spurious, such as Tooth enamel or Transmission Control Protocol. Since the process of feature generation relies on the bag of words for matching concepts to the input text, it suffers from the BOW shortcomings we mentioned above (Section 4.1). Consequently, some features are generated because the corresponding Wikipedia articles just happen to share words with the input text, even though these words are not characteristic of the article as a whole. As explained above, our method can successfully operate in the presence of such extraneous features due to the use of feature selection. This way, generated features that are not informative for predicting document categories are filtered out, and only informative features are actually retained for learning the classification model.

**Using Inter-article Links for Generating Additional Features** In Section 1, we presented an algorithm that generates additional features using inter-article links as relations between concepts. In what follows, we show a series of text fragments, where for each fragment we show (a) features generated with the regular FG algorithm, (b) features generated using Wikipedia links, and (c) more general features generated using links. As we can see from the examples, the features constructed using the links are often relevant to the input text.

- **Text:** *"Google search"*

  **Regular feature generation:** (1) Search engine; (2) Google Video; (3) Google; (4) Google (search); (5) Google Maps; (6) Google Desktop; (7) Google (verb); (8) Google News; (9) Search engine optimization; (10) Spamdexing (search engine spamming)

  **Features generated using links:** (1) PageRank; (2) AdWords; (3) AdSense; (4) Gmail; (5) Google Platform; (6) Website; (7) Sergey Brin; (8) Google bomb; (9) MSN Search; (10) Nigritude ultramarine (a meaningless phrase used in a search engine optimization contest in 2004)

  **More general features only:** (1) Website; (2) Mozilla Firefox; (3) Portable Document Format; (4) Algorithm; (5) World Wide Web

- **Text:** *"programming tools"*

  **Regular feature generation:** (1) Tool; (2) Programming tool; (3) Computer software; (4) Integrated development environment; (5) Computer-aided software engineering; (6) Macromedia Flash; (7) Borland; (8) Game programmer; (9) C programming language; (10) Performance analysis

  **Features generated using links:** (1) Compiler; (2) Debugger; (3) Source code; (4) Software engineering; (5) Microsoft; (6) Revision control; (7) Scripting language; (8) GNU; (9) Make; (10) Linux

  **More general features only:** (1) Microsoft; (2) Software engineering; (3) Linux; (4) Compiler; (5) GNU

### 4.5.2 The Effect of Feature Generation

Table 4 shows the results of using Wikipedia-based feature generation, with significant improvements ($p < 0.05$) shown in bold. The different rows of the table correspond to the performance on different datasets and their subsets, as defined in Section 4.3. We consistently observed larger improvements in macro-averaged BEP, which is dominated by categorization effectiveness on small categories. This goes in line with our expectations that the contribution of encyclopedic knowledge should be especially prominent for categories with few training examples. Categorization performance was improved for virtually all datasets, with notable improvements of up to 30.4% for RCV1 and 18% for OHSUMED. Using the Wilcoxon test, we found that the Wikipedia-based classifier is significantly superior to the baseline with $p < 10^{-5}$ in both micro- and macro-averaged cases. These results clearly demonstrate the advantage of knowledge-based feature generation.

In our prior work (Gabrilovich & Markovitch, 2005, 2007b), we have also performed feature generation for text categorization using an alternative source of knowledge, namely, the Open Directory Project (ODP). The results of using Wikipedia are competitive with those using ODP, with a slight advantage of Wikipedia. Observe also that Wikipedia is constantly updated by numerous volunteers around the globe, while the ODP is virtually frozen nowadays. Hence, in the future we can expect to obtain further improvements by using newer versions of Wikipedia.

**The Effect of Knowledge Breadth** We also examined the effect of performing feature generation using a newer Wikipedia snapshot, as explained in Section 3.2.2. Appendix A reports the results of this experiment, which show a small but consistent improvement due to using a larger knowledge base.

### 4.5.3 Classifying Short Documents

We conjectured that Wikipedia-based feature generation should be particularly useful for classifying short documents.

Table 5 presents the results of this evaluation on the datasets defined in Section 4.3. In the majority of cases, feature generation yielded greater improvement on short documents than on regular documents. Notably, the improvements are particularly high for OHSUMED, where "pure" experimentation on short documents is possible (see Section 4.3). According to the Wilcoxon test, the Wikipedia-based classifier is significantly superior to the baseline with $p < 2 \cdot 10^{-6}$. These findings confirm our hypothesis that encyclopedic knowledge should be particularly useful when categorizing short documents, which are inadequately represented by the standard bag of words.

### 4.5.4 Using Inter-article links as Concept Relations

Using inter-article links for generating additional features, we observed further improvements in text categorization performance on short documents. As we can see in Table 6, in the absolute majority of cases using links to generate more general features only is a superior strategy. As we explain in Section 2.3, inter-article links can be viewed as relations between concepts represented by the articles. Consequently, using these links allows us to identify additional concepts related to the context being analyzed, which leads to better representation of the context with additional relevant generated features.

| Dataset | Baseline | | Wikipedia | | Improvement | |
|---|---|---|---|---|---|---|
| | micro | macro | micro | macro | micro | macro |
| | BEP | BEP | BEP | BEP | BEP | BEP |
| Reuters-21578 (10 cat.) | 0.925 | 0.874 | 0.932 | 0.887 | +0.8% | +1.5% |
| Reuters-21578 (90 cat.) | 0.877 | 0.602 | 0.883 | 0.603 | +0.7% | +0.2% |
| RCV1 Industry-16 | 0.642 | 0.595 | 0.645 | **0.617** | +0.5% | **+3.7%** |
| RCV1 Industry-10A | 0.421 | 0.335 | **0.448** | **0.437** | **+6.4%** | **+30.4%** |
| RCV1 Industry-10B | 0.489 | 0.528 | **0.523** | **0.566** | **+7.0%** | **+7.2%** |
| RCV1 Industry-10C | 0.443 | 0.414 | **0.468** | **0.431** | **+5.6%** | **+4.1%** |
| RCV1 Industry-10D | 0.587 | 0.466 | 0.595 | 0.459 | +1.4% | -1.5% |
| RCV1 Industry-10E | 0.648 | 0.605 | 0.641 | 0.612 | -1.1% | +1.2% |
| RCV1 Topic-16 | 0.836 | 0.591 | 0.843 | **0.661** | +0.8% | **+11.8%** |
| RCV1 Topic-10A | 0.796 | 0.587 | 0.798 | **0.682** | +0.3% | **+16.2%** |
| RCV1 Topic-10B | 0.716 | 0.618 | 0.723 | **0.656** | +1.0% | **+6.1%** |
| RCV1 Topic-10C | 0.687 | 0.604 | 0.699 | 0.618 | +1.7% | +2.3% |
| RCV1 Topic-10D | 0.829 | 0.673 | 0.839 | 0.688 | +1.2% | +2.2% |
| RCV1 Topic-10E | 0.758 | 0.742 | 0.765 | 0.755 | +0.9% | +1.8% |
| OHSUMED-10A | 0.518 | 0.417 | **0.538** | **0.492** | **+3.9%** | **+18.0%** |
| OHSUMED-10B | 0.656 | 0.500 | 0.667 | **0.534** | +1.7% | **+6.8%** |
| OHSUMED-10C | 0.539 | 0.505 | 0.545 | **0.522** | +1.1% | **+3.4%** |
| OHSUMED-10D | 0.683 | 0.515 | 0.692 | **0.546** | +1.3% | **+6.0%** |
| OHSUMED-10E | 0.442 | 0.542 | **0.462** | **0.575** | **+4.5%** | **+6.1%** |
| 20NG | 0.854 | | **0.862** | | **+1.0%** | |
| Movies | 0.813 | | **0.842** | | **+3.6%** | |

Table 4: The effect of feature generation for long documents

| Dataset | Baseline | | Wikipedia | | Improvement | |
|---|---|---|---|---|---|---|
| | micro | macro | micro | macro | micro | macro |
| | BEP | BEP | BEP | BEP | BEP | BEP |
| Reuters-21578 (10 cat.) | 0.868 | 0.774 | 0.877 | 0.793 | +1.0% | +2.5% |
| Reuters-21578 (90 cat.) | 0.793 | 0.479 | 0.803 | **0.506** | +1.3% | **+5.6%** |
| RCV1 Industry-16 | 0.454 | 0.400 | **0.481** | **0.437** | **+5.9%** | **+9.2%** |
| RCV1 Industry-10A | 0.249 | 0.199 | **0.293** | **0.256** | **+17.7%** | **+28.6%** |
| RCV1 Industry-10B | 0.273 | 0.292 | **0.337** | **0.363** | **+23.4%** | **+24.3%** |
| RCV1 Industry-10C | 0.209 | 0.199 | **0.294** | **0.327** | **+40.7%** | **+64.3%** |
| RCV1 Industry-10D | 0.408 | 0.361 | **0.452** | **0.379** | **+10.8%** | **+5.0%** |
| RCV1 Industry-10E | 0.450 | 0.410 | **0.474** | **0.434** | **+5.3%** | **+5.9%** |
| RCV1 Topic-16 | 0.763 | 0.529 | 0.769 | 0.542 | +0.8% | +2.5% |
| RCV1 Topic-10A | 0.718 | 0.507 | 0.725 | **0.544** | +1.0% | **+7.3%** |
| RCV1 Topic-10B | 0.647 | 0.560 | 0.643 | 0.564 | -0.6% | +0.7% |
| RCV1 Topic-10C | 0.551 | 0.471 | **0.573** | **0.507** | **+4.0%** | **+7.6%** |
| RCV1 Topic-10D | 0.729 | 0.535 | 0.735 | **0.563** | +0.8% | **+5.2%** |
| RCV1 Topic-10E | 0.643 | 0.636 | 0.670 | 0.653 | +4.2% | +2.7% |
| OHSUMED-10A | 0.302 | 0.221 | **0.405** | **0.299** | **+34.1%** | **+35.3%** |
| OHSUMED-10B | 0.306 | 0.187 | **0.383** | **0.256** | **+25.2%** | **+36.9%** |
| OHSUMED-10C | 0.441 | 0.296 | **0.528** | **0.413** | **+19.7%** | **+39.5%** |
| OHSUMED-10D | 0.441 | 0.356 | **0.460** | **0.402** | **+4.3%** | **+12.9%** |
| OHSUMED-10E | 0.164 | 0.206 | **0.219** | **0.280** | **+33.5%** | **+35.9%** |
| 20NG | 0.699 | | **0.749** | | **+7.1%** | |

Table 5: Feature generation for short documents

| Dataset | Baseline | | Wikipedia | | Wikipedia + links | | Wikipedia + links (more general features only) | |
|---|---|---|---|---|---|---|---|---|
| | micro BEP | macro BEP | micro BEP | macro BEP | micro BEP | macro BEP | micro BEP | macro BEP |
| Reuters-21578 (10 cat.) | 0.868 | 0.774 | 0.877 | 0.793 | 0.878 | 0.796 | 0.880 | 0.801 |
| Reuters-21578 (90 cat.) | 0.793 | 0.479 | 0.803 | 0.506 | 0.804 | 0.506 | 0.809 | 0.507 |
| RCV1 Industry-16 | 0.454 | 0.400 | 0.481 | 0.437 | 0.486 | 0.445 | 0.488 | 0.444 |
| RCV1 Topic-16 | 0.763 | 0.529 | 0.769 | 0.542 | 0.769 | 0.539 | 0.775 | 0.545 |
| 20NG | 0.699 | | 0.749 | | 0.753 | | 0.756 | |
| Dataset | | | Improvement over baseline | | Improvement over baseline | | Improvement over baseline | |
| Reuters-21578 (10 cat.) | – | – | +1.0% | +2.5% | +1.2% | +2.8% | +1.4% | +3.5% |
| Reuters-21578 (90 cat.) | – | – | +1.3% | +5.6% | +1.4% | +5.6% | +2.0% | +5.8% |
| RCV1 Industry-16 | – | – | +5.9% | +9.2% | +7.1% | +11.3% | +7.5% | +11.0% |
| RCV1 Topic-16 | – | – | +0.8% | +2.5% | +0.8% | +1.9% | +1.6% | +3.0% |
| 20NG | – | | +7.1% | | +7.7% | | +8.1% | |

Table 6: Feature generation for short documents using inter-article links

## 5. Related Work

This section puts our methodology in the context of related prior work.

In the past, there have been a number of attempts to represent the meaning of natural language texts. Early research in computational linguistics focused on deep natural language understanding, and strived to represent text semantics using logical formulae (Montague, 1973). However, this task proved to be very difficult and little progress has been made to develop comprehensive grammars for non-trivial fragments of the language. Consequently, the mainstream research effectively switched to more statistically-based methods (Manning & Schuetze, 2000).

Although few of these studies tried to explicitly define semantic representation, their *modus operandi* frequently induces a particular representation system. Distributional similarity methods (Lee, 1999) compute the similarity of a pair of words $w_1$ and $w_2$ by comparing the distributions of other words given these two, e.g., by comparing vectors of probabilities $P(v|w_1)$ and $P(v|w_2)$ for a large vocabulary $V$ of words ($v \in V$). Therefore, these techniques can be seen as representing the meaning of a word $w$ as a vector of conditional probabilities of other words given $w$. Dagan, Marcus, and Markovitch (1995) refined this technique by considering co-occurrence probabilities of a word with its left and right contextual neighbors. For example, the word "water" would be represented by a vector of its left neighbors such as "drink", "pour", and "clean", and the vector of right neighbors such as "molecule", "level", and "surface". Lin (1998a) represented word meaning by considering syntactic roles of other words that co-occur with it in a sentence. For example, the semantics of the word "water" would be represented by a vector of triples such as (`water, obj-of, drink`) and (`water, adj-mod, clean`). Qiu and Frei (1993) proposed a method for concept-based query expansion; however, they expanded queries with additional words rather than with features corresponding to semantic concepts.

Latent Semantic Analysis is probably the most similar method in prior research, as it does explicitly represents the meaning of a text fragment. LSA does so by manipulating a vector of so-called latent concepts, which are obtained through SVD decomposition of a word-by-document matrix of the training corpus. CYC (Lenat, 1995; Lenat, Guha, Pittman, Pratt, & Shepherd, 1990) represents semantics of words through an elaborate network of interconnected and richly-annotated concepts.

In contrast, our method represents the meaning of a piece of text as a weighted vector of knowledge concepts. Importantly, entries of this vector correspond to unambiguous human-defined concepts rather than plain words, which are often ambiguous. Compared to LSA, our approach benefits from large amounts of manually encoded human knowledge, as opposed to defining concepts using statistical analysis of a training corpus. Compared to CYC, our approach streamlines the process of semantic interpretation that does not depend on manual encoding of inference rules. With the exception of LSA, most prior approaches to semantic interpretation explicitly represent semantics of individual words, and require an extra level of sophistication to represent longer texts. Conversely, our approach represents the meaning of texts in a uniform way regardless of their length.

## 5.1 Semantic Similarity and Semantic Relatedness

In this study we deal with "semantic relatedness" rather than "semantic similarity" or "semantic distance", which are also often used in the literature. In their extensive survey of relatedness measures, Budanitsky and Hirst (2006) argued that the notion of relatedness is more general than that of similarity, as the former subsumes many different kind of specific relations, including meronymy, antonymy, functional association, and others. They further maintained that computational linguistics applications often require measures of relatedness rather than the more narrowly defined measures of similarity. For example, word sense disambiguation can use any *related* words from the context, and not merely *similar* words. Budanitsky and Hirst (2006) also argued that the notion of semantic distance might be confusing due to the different ways it has been used in the literature.

Our approach to estimating semantic relatedness of words is somewhat reminiscent of distributional (or co-occurrence) similarity (Lee, 1999; Dagan, Lee, & Pereira, 1999). Indeed, we compare the meanings of words by comparing the occurrence patterns across a large collection of natural language documents. However, the compilation of these documents is not arbitrary, rather, the documents are aligned with encyclopedia articles, while each of them is focused on a single topic. Furthermore, distributional similarity methods are inherently suitable for comparing individual words, while our method can compute similarity of arbitrarily long texts.

Prior work in the field mostly focused on semantic *similarity* of words, using R&G (Rubenstein & Goodenough, 1965) list of 65 word pairs and M&C (Miller & Charles, 1991) list of 30 word pairs. When only the similarity relation is considered, using lexical resources was often successful enough, reaching the Pearson's correlation of 0.70–0.85 with human judgements (Budanitsky & Hirst, 2006; Jarmasz, 2003). In this case, lexical techniques even have a slight edge over ESA-Wikipedia, whose correlation with human scores is 0.723 on M&C and 0.816 on R&G. [21] However, when the entire language wealth is considered in an attempt to capture more general semantic relatedness, lexical techniques yield substantially inferior results (see Table 2). WordNet-based technique, which only consider the generalization ("is-a") relation between words, achieve correlation of only 0.33–0.35 with human judgements (Budanitsky & Hirst, 2006; Jarmasz, 2003). Jarmasz & Szpakowicz's ELKB system (Jarmasz, 2003) based on Roget's Thesaurus (Roget, 1852) achieves a higher correlation of 0.55 due to its use of a richer set if relations.

Studying semantic similarity and relatedness of words is related to assessing similarity of relations. An example of this task is to establish that word pairs *carpenter:wood* and *mason:stone* are relationally similar, as the words in both pairs stand in the same relation (profession:material). State of the art results on relational similarity are based on Latent Relational Analysis (Turney, 2006, 2005).

Sahami and Heilman (2006) proposed to use the Web as a source of additional knowledge for measuring similarity of short text snippets. To this end, they defined a kernel function that sends two snippets as queries to a search engine, and compares the bags of words for the two sets of returned documents. A major limitation of this technique is that it is only applicable to short texts, because sending a long text as a query to a search engine is likely to return few or even no results at all. On the other hand, our approach is applicable to text fragments of arbitrary length. Additional studies that explored the Web to gather information for computing word similarity include (Turney, 2001) and (Metzler, Dumais, & Meek, 2007). The main difference between these works and our method is that the latter uses a structured representation of human knowledge defined by Wikipedia concepts.

The above-mentioned based techniques are inherently limited to individual words, and their adaptation for comparing longer texts requires an extra level of complexity (Mihalcea et al., 2006). In contrast, our method treats both words and texts in essentially the same way.

Strube and Ponzetto (2006) also used Wikipedia for computing semantic relatedness. However, their method, called WikiRelate!, is radically different from ours. Given a pair of words $w_1$ and $w_2$, WikiRelate! searches for Wikipedia articles, $p_1$ and $p_2$, that respectively contain $w_1$ and $w_2$ in their titles. Semantic relatedness is then computed based on various distance measures between $p_1$ and $p_2$. These measures either rely on the texts of the pages, or path distances within the category hierarchy of Wikipedia. Our approach, on the other hand, represents each word as a weighted vector of Wikipedia concepts. Semantic relatedness is then computed by comparing the two concept vectors.

Thus, the differences between the two approaches are:

1. WikiRelate! can only process words that actually occur in titles of Wikipedia articles. ESA only requires that the word appears within the text of Wikipedia articles.

2. WikiRelate! is limited to single words while ESA can compare texts of any length.

---

21. WikiRelate! (Strube & Ponzetto, 2006) achieved relatively low scores of 0.31–0.54 on these domains.

3. WikiRelate! represents the semantics of a word by either the text of the article associated with it, or by the node in the category hierarchy. ESA has a much more structured semantic representation consisting of a vector of Wikipedia concepts.

Indeed, as we have shown in Section 3.3, the richer representation of ESA yields much better results.

## 5.2 Feature Generation for Text Categorization

To date, quite a few attempts have been made to deviate from the orthodox bag of words paradigm, usually with limited success. In particular, representations based on phrases (Lewis, 1992; Dumais et al., 1998; Fuernkranz, Mitchell, & Riloff, 1998), named entities (Kumaran & Allan, 2004), and term clustering (Lewis & Croft, 1990; Bekkerman, 2003) have been explored. However, none of these techniques could possibly overcome the problem underlying the various examples we reviewed in this paper—lack of world knowledge.

*Feature generation* techniques were found useful in a variety of machine learning tasks (Markovitch & Rosenstein, 2002; Fawcett, 1993; Matheus, 1991). These techniques search for new features that describe the target concept better than the ones supplied with the training instances. A number of proposed feature generation algorithms (Pagallo & Haussler, 1990; Matheus & Rendell, 1989; Hu & Kibler, 1996; Murphy & Pazzani, 1991; Hirsh & Japkowicz, 1994) led to significant improvements in performance over a range of classification tasks. However, even though feature generation is an established research area in machine learning, only a few works have applied it to text processing (Kudenko & Hirsh, 1998; Mikheev, 1998; Cohen, 2000; Scott, 1998; Scott & Matwin, 1999). In contrast to our approach, these techniques did not use any exogenous knowledge.

In our prior work (Gabrilovich & Markovitch, 2005, 2007b), we assumed the external knowledge is available in the form of a generalization hierarchy, and used the Open Directory Project as an example. This method, however, had a number of drawbacks, which can be corrected by using Wikipedia.

First, requiring the knowledge repository to define an "is-a" hierarchy limits the choice of appropriate repositories. Moreover, hierarchical organization embodies only one particular relation between the nodes (generalization), while numerous other relations, such as relatedness, meronymy/holonymy and chronology, are ignored. Second, large-scale hierarchies tend to be extremely unbalanced, so that the relative size of some branches is disproportionately large or small due to peculiar views of the editors. Such phenomena are indeed common in the ODP. For example, the TOP/SOCIETY branch is heavily dominated by one of its children—RELIGION AND SPIRITUALITY; the TOP/SCIENCE branch is dominated by its BIOLOGY child; a considerable fraction of the mass of TOP/RECREATION is concentrated in PETS. Finally, to learn the scope of every ODP concept, short textual descriptions of the concepts were augmented by crawling the Web sites cataloged in the ODP. This procedure allowed us to accumulate many gigabytes worth of textual data, but at a price, as texts obtained from the Web are often quite far from formal writing and plagued with noise. Crawling a typical Web site often brings auxiliary material that has little to do with the site theme, such as legal disclaimers, privacy statements, and help pages.

In this paper we proposed to use world knowledge encoded in Wikipedia, which is arguably the largest knowledge repository on the Web. Compared to the ODP, Wikipedia possesses several advantageous properties. First, its articles are much cleaner than typical Web pages, and mostly qualify as standard written English. Although Wikipedia offers several orthogonal browsing interfaces, their structure is fairly shallow, and we propose to treat Wikipedia as having essentially no hierarchy. This way, mapping tex fragments onto relevant Wikipedia concepts yields truly multi-faceted classification of the text, and avoids the problem of unbalanced hierarchy branches. Moreover, by not requiring the knowledge repository to be hierarchically organized, our approach is suitable for new domains, for which no ontology is available. Finally, Wikipedia articles are heavily cross-linked, in a way reminiscent of linking on the Web. We conjectured that these links encode many interesting relations between the concepts, and constitute an important source of information in addition to the article texts. We explored using inter-article links in Section 4.5.4.

### 5.2.1 Feature Generation Using Electronic Dictionaries

Several studies performed feature construction using the WordNet electronic dictionary (Fellbaum, 1998) and other domain-specific dictionaries (Scott, 1998; Scott & Matwin, 1999; Urena-Lopez, Buenaga, & Gomez, 2001; Wang, McKay, Abbass, & Barlow, 2003; Bloehdorn & Hotho, 2004).

Scott and Matwin (1999) attempted to augment the conventional bag-of-words representation with additional features, using the symbolic classification system Ripper (Cohen, 1995). This study evaluated features based on syntactically[22] and statistically motivated phrases, as well as on WordNet *synsets*[23]. In the latter case, the system performed generalizations using the hypernym hierarchy of WordNet, and completely replaced a bag of words with a bag of synsets. While using hypernyms allowed Ripper to produce more general and more comprehensible rules and achieved some performance gains on small classification tasks, no performance benefits could be obtained for larger tasks, which even suffered from some degradation in classification accuracy. Consistent with other published findings (Lewis, 1992; Dumais et al., 1998; Fuernkranz et al., 1998), the phrase-based representation also did not yield any significant performance benefits over the bag-of-words approach.[24]

Urena-Lopez et al. (2001) used WordNet in conjunction with Rocchio (Rocchio, 1971) and Widrow-Hoff (Lewis, Schapire, Callan, & Papka, 1996; Widrow & Stearns, 1985, Chapter 6) linear classifiers to fine-tune the category vectors. Wang et al. (2003) used Medical Subject Headings (MeSH, 2003) to replace the bag of words with canonical medical terms; Bloehdorn and Hotho (2004) used a similar approach to augment Reuters-21578 documents with WordNet synsets and OHSUMED medical documents with MeSH terms.

It should be noted, however, that WordNet was not originally designed to be a powerful knowledge base, but rather a lexical database more suitable for peculiar lexicographers' needs. Specifically, WordNet has the following drawbacks when used as a knowledge base for text categorization:

---

22. Identification of syntactic phrases was performed using a noun phrase extractor built on top of a part of speech tagger (Brill, 1995).

23. A *synset* is WordNet notion for a sense shared by a group of synonymous words.

24. Sebastiani (2002) casts the use of bag of words versus phrases as utilizing *lexical semantics* rather than *compositional semantics*. Interestingly, some bag-of-words approaches (notably, KNN) may be considered *context-sensitive* as they do not assume independence between either features (terms) or categories (Yang & Pedersen, 1997).

- WordNet has a fairly small coverage—for the test collections we used in this paper, up to 50% of their unique words are missing from WordNet. In particular, many proper names, slang and domain-specific technical terms are not included in WordNet, which was designed as a general-purpose dictionary.

- Additional information about synsets (beyond their identity) is very limited. This is because WordNet implements a *differential* rather than *constructive* lexical semantics theory, so that glosses that accompany the synsets are mainly designed to distinguish the synsets rather than provide a definition of the sense or concept. Usage examples that occasionally constitute part of the gloss serve the same purpose. Without such auxiliary information, reliable word sense disambiguation is almost impossible.

- WordNet was designed by professional linguists who are trained to recognize minute differences in word senses. As a result, common words have far too many distinct senses to be useful in information retrieval (Mihalcea, 2003); for example, the word "make" has as many as 48 senses as a verb alone. Such fine-grained distinctions between synsets present an additional difficulty for word sense disambiguation.

Both our approach and the techniques that use WordNet manipulate a collection of concepts. However, there are a number of crucial differences. All previous studies only performed feature generation for individual words only. Our approach can handle arbitrarily long or short text fragments alike. Considering words in context allows our approach to perform word sense disambiguation. Approaches using WordNet cannot achieve disambiguation because information about synsets is limited to merely a few words, while in Wikipedia concepts are associated with huge amounts of text. Even for individual words, our approach provides much more sophisticated mapping of words to concepts, through the analysis of the large bodies of texts associated with concepts. This allows us to represent the meaning of words (or texts) as a weighted combination of concepts, while mapping a word in WordNet amounts to simple lookup, without any weights. Furthermore, in WordNet the senses of each word are mutually exclusive. In our approach, concepts reflect different aspects of the input, thus yielding weighted multi-faceted representation of the text.

In Appendix D we illustrate the limitations of WordNet on a specific example, where we juxtapose WordNet-based and Wikipedia-based representation.

### 5.2.2 USING UNLABELED EXAMPLES

To the best of our knowledge, with the exception of the above studies that used WordNet, there have been no attempts to date to automatically use large-scale repositories of structured background knowledge for feature generation. An interesting approach to using non-structured background knowledge was proposed by Zelikovitz and Hirsh (2000). This work uses a collection of unlabeled examples as intermediaries in comparing testing examples with the training ones. Specifically, when an unknown test instance does not appear to resemble any labeled training instances, unlabeled examples that are similar to both may be used as "bridges." Using this approach, it is possible to handle the situation where the training and the test document have few or no words in common. The unlabeled documents are utilized to define a cosine similarity metric, which is then used by the KNN algorithm for actual text categorization. This approach, however, suffers from efficiency problems, as looking for intermediaries to compare every two documents makes it necessary to explore a combinatorial search space.

In a subsequent paper, Zelikovitz and Hirsh (2001) proposed an alternative way to use unlabeled documents as background knowledge. In this work, unlabeled texts are pooled together with the training documents to compute a Latent Semantic Analysis (LSA) (Deerwester et al., 1990) model. LSA analyzes a large corpus of unlabeled text, and automatically identifies so-called "latent concepts" using Singular Value Decomposition. The resulting LSA metric then facilitates comparison of test documents to training documents. The addition of unlabeled documents significantly increases the amount of data on which word co-occurrence statistics is estimated, thus providing a solution to text categorization problems where training data is particularly scarce. However, subsequent studies found that LSA can rarely improve the strong baseline established by SVM, and often even results in performance degradation (Wu & Gunopulos, 2002; Liu, Chen, Zhang, Ma, & Wu, 2004). In contrast to LSA, which manipulates virtual concepts, our methodology relies on using concepts identified and described by humans.

## 6. Conclusions

In this paper we proposed Explicit Semantic Analysis—a semantic interpretation methodology for natural language processing. In order to render computers with knowledge about the world, we use Wikipedia to build a semantic interpreter, which represents the meaning of texts in a very high-dimensional space of knowledge-based concepts. These concepts correspond to Wikipedia articles, and our methodology provides a fully automatic way to tap into the collective knowledge of tens and hundreds of thousands of people. The concept-based representation of text contains information that cannot be deduced from the input text alone, and consequently supersedes the conventional bag of words representation.

We believe the most important aspects of the proposed approach are its ability to address synonymy and polysemy, which are arguably the two most important problems in NLP. Thus, the two texts can discuss the same topic using different words, and the conventional bag of words approach will not be able to identify this commonality. On the other hand, the mere fact that the two texts contain the same word does not necessarily imply that they discuss the same topic, since that word could be used in the two texts in two different meanings. We believe that our concept-based representation allows generalizations and refinements to partially address synonymy and polysemy.

Consider, for example, the following text fragment (taken from Appendix C): "A group of European-led astronomers has made a photograph of what appears to be a planet orbiting another star. If so, it would be the first confirmed picture of a world beyond our solar system." The fifth concept generated for this fragment is Extrasolar planet, which is exactly the topic of this text, even though these words are not mentioned in the input. The other generated concepts (e.g., Astronomy and Planetary orbit) are also highly characteristic of astronomy-related texts. Such additions enrich the text representation, and increase the chances of finding common features between texts. It is also essential to note that, of course, not all the generated concepts need to match features of other documents. Even if some of the concepts match, we gain valuable insights about the document contents.

We succeeded to make automatic use of an encyclopedia without deep language understanding, specially crafted inference rules or relying on additional common-sense knowledge bases. This was made possible by applying standard text classification techniques to match document texts with relevant Wikipedia articles.

Empirical evaluation confirmed the value of Explicit Semantic Analysis for two common tasks in natural language processing. Compared with the previous state of the art, using ESA results in significant improvements in automatically assessing semantic relatedness of words and texts. Specifically, the correlation of computed relatedness scores with human judgements increased from $r = 0.56$ to $0.75$ (Spearman) for individual words and from $r = 0.60$ to $0.72$ (Pearson) for texts. In contrast to existing methods, ESA offers a uniform way for computing relatedness of both individual words and arbitrarily long text fragments. Using ESA to perform feature generation for text categorization yielded consistent improvements across a diverse range of datasets. Recently, the performance of the best text categorization systems became similar, and previous work mostly achieved small improvements. Using Wikipedia as a source of external knowledge allowed us to improve the performance of text categorization across a diverse collection of datasets.

It should be noted that although a recent study (Giles, 2005) found Wikipedia accuracy to rival that of Encyclopaedia Britannica, arguably not all the Wikipedia articles are of equally high quality. On the one hand, Wikipedia has the notion of featured articles (`http://en.wikipedia.org/wiki/Featured_Article`), which "are considered to be the best articles in Wikipedia, as determined by Wikipedia's editors." Currently, fewer than 0.1% of articles achieve this status. On the other hand, many articles are incomplete (so-called stubs), or might even contain information that is incorrect or that does not represent a consensus among the editors. Yet in other cases, Wikipedia content might be prone to spamming, despite the editorial process that attempts to review recent changes. We believe our method is not overly susceptible to such cases, as long as the majority of the content is correct. Arguably, except for outright vandalism, most spamming would likely modify articles to contain information that is related to the topic of the article, but not important or not essential for the majority of readers. As long as this newly added content remains relevant to the gist of the article, our method will likely be able to correctly determine those input texts that the article is relevant for. However, a proper evaluation of the robustness of our method in the presence of imperfect content is beyond the scope of this article.

We believe that this research constitutes a step towards enriching natural language processing with humans' knowledge about the world. We hope that Explicit Semantic Analysis will also be useful for other NLP tasks beyond computing semantic relatedness and text categorization, and we intend to investigate this in our future work. Recently, we have used ESA to improve the performance of conventional information retrieval (Egozi, Gabrilovich, & Markovitch, 2008). In that work, we augmented both queries and documents with generated features, such that documents were indexed in the augmented space of words and concepts. Potthast, Stein, and Anderka (2008) and Sorg and Cimiano (2008) adapted ESA for multi-lingual and cross-lingual information retrieval.

In another recent study, Gurevych et al. (2007) applied our methodology to computing word similarity in German, and also to an information retrieval task that searched job descriptions given a user's description of her career interests, and found our method superior to a WordNet-based approach. Importantly, this study also confirms that our method can be easily adapted to languages other than English, by using the version of Wikipedia corresponding to the desired target language.

In our future work, we also intend to apply ESA to word sense disambiguation. Current approaches to word sense disambiguation represent contexts that contain ambiguous words using the bag of words augmented with part-of-speech information. We believe representation of such contexts can be greatly improved if we use feature generation to map such contexts into relevant knowledge concepts. Anecdotal evidence (such as the examples presented in Section 4.5.1) implies our method has promise for improving the state of the art in word sense disambiguation. In this work we capitalized on inter-article links of Wikipedia in several ways, and in our future work we intend to investigate more elaborate techniques for leveraging the high degree of cross-linking between Wikipedia articles.

The Wiki technology underlying the Wikipedia project is often used nowadays in a variety of open-editing initiatives. These include corporate intranets that use Wiki as a primary documentation tool, as well as numerous domain-specific encyclopedias on topics ranging from mathematics to Orthodox Christianity.[25] Therefore, we believe our methodology can also be used for augmenting document representation in many specialized domains. It is also essential to note that Wikipedia is available in numerous languages, while different language versions are cross-linked at the level of concepts. We believe this information can be leveraged to use Wikipedia-based semantic interpretation for improving machine translation.

This work proposes a methodology for Explicit Semantic Analysis using Wikipedia. However, ESA can also be implemented using other repositories of human knowledge that satisfy the requirements listed in Section 2.1. In Section 3.3 we reported the results of building an ESA-based semantic interpreter using the Open Directory Project (Gabrilovich & Markovitch, 2005, 2007b). Zesch, Mueller, and Gurevych (2008) proposed to use Wiktionary for computing semantic relatedness. In our future work, we intend to implement ESA using additional knowledge repositories.

Finally, for readers interested in using Wikipedia in their own work, the main software deliverable of the described work is the Wikipedia preprocessor (WikiPrep), available online as part of the SourceForge open-source project at `http://wikiprep.sourceforge.net`.

## Acknowledgments

We thank Michael D. Lee and Brandon Pincombe for making available their document similarity data. We also thank Deepak Agarwal for advice on assessing statistical significance of results in computing semantic relatedness. This work was partially supported by funding from the EC-sponsored MUSCLE Network of Excellence.

The first author's current address is Yahoo! Research, 2821 Mission College Blvd, Santa Clara, CA 95054, USA.

---

25. See `http://en.wikipedia.org/wiki/Category:Online_encyclopedias` for a longer list of examples.

## Appendix A. The effect of knowledge breadth in text categorization

In this appendix, we examine the effect of performing feature generation using a newer Wikipedia snapshot, as defined in Section 3.2.2. As we can see from Table 7, using the larger amount of knowledge leads on average to greater improvements in text categorization performance. Although the difference between the performance of the two versions is admittedly small, it is consistent across datasets (a similar situation happens when assessing the role of external knowledge for computing semantic relatedness, see Section 3.3.3).

| Dataset | Baseline | | Wikipedia (26/03/06) | | Improvement (26/03/06) | | Improvement (05/11/05) | |
|---|---|---|---|---|---|---|---|---|
| | micro BEP | macro BEP | micro BEP | macro BEP | micro BEP | macro BEP | micro BEP | macro BEP |
| Reuters-21578 (10 cat.) | 0.925 | 0.874 | 0.935 | 0.891 | +1.1% | +1.9% | +0.8% | +1.5% |
| Reuters-21578 (90 cat.) | 0.877 | 0.602 | 0.883 | 0.600 | +0.7% | -0.3% | +0.7% | +0.2% |
| RCV1 Industry-16 | 0.642 | 0.595 | 0.648 | **0.616** | +0.9% | **+3.5%** | +0.5% | **+3.7%** |
| RCV1 Industry-10A | 0.421 | 0.335 | **0.457** | **0.450** | **+8.6%** | **+34.3%** | **+6.4%** | **+30.4%** |
| RCV1 Industry-10B | 0.489 | 0.528 | **0.527** | **0.559** | **+7.8%** | **+5.9%** | **+7.0%** | **+7.2%** |
| RCV1 Industry-10C | 0.443 | 0.414 | **0.458** | 0.424 | **+3.4%** | +2.4% | **+5.6%** | **+4.1%** |
| RCV1 Industry-10D | 0.587 | 0.466 | 0.607 | **0.448** | +3.4% | **-3.9%** | +1.4% | -1.5% |
| RCV1 Industry-10E | 0.648 | 0.605 | 0.649 | 0.607 | +0.2% | +0.3% | -1.1% | +1.2% |
| RCV1 Topic-16 | 0.836 | 0.591 | 0.842 | **0.659** | +0.7% | **+11.5%** | +0.8% | **+11.8%** |
| RCV1 Topic-10A | 0.796 | 0.587 | 0.802 | **0.689** | +0.8% | **+17.4%** | +0.3% | **+16.2%** |
| RCV1 Topic-10B | 0.716 | 0.618 | 0.725 | **0.660** | +1.3% | **+6.8%** | +1.0% | **+6.1%** |
| RCV1 Topic-10C | 0.687 | 0.604 | 0.697 | **0.627** | +1.5% | **+3.8%** | +1.7% | +2.3% |
| RCV1 Topic-10D | 0.829 | 0.673 | 0.838 | 0.687 | +1.1% | +2.1% | +1.2% | +2.2% |
| RCV1 Topic-10E | 0.758 | 0.742 | 0.762 | 0.752 | +0.5% | +1.3% | +0.9% | +1.8% |
| OHSUMED-10A | 0.518 | 0.417 | **0.545** | **0.490** | **+5.2%** | **+17.5%** | **+3.9%** | **+18.0%** |
| OHSUMED-10B | 0.656 | 0.500 | 0.667 | **0.529** | +1.7% | **+5.8%** | +1.7% | **+6.8%** |
| OHSUMED-10C | 0.539 | 0.505 | 0.553 | **0.527** | +2.6% | **+4.4%** | +1.1% | **+3.4%** |
| OHSUMED-10D | 0.683 | 0.515 | 0.694 | **0.550** | +1.6% | **+6.8%** | +1.3% | **+6.0%** |
| OHSUMED-10E | 0.442 | 0.542 | **0.461** | **0.588** | **+4.3%** | **+8.5%** | **+4.5%** | **+6.1%** |
| 20NG | 0.854 | | **0.859** | | **+0.6%** | | **+1.0%** | |
| Movies | 0.813 | | **0.850** | | **+4.5%** | | **+3.6%** | |
| *Average* | | | | | *+2.50%* | *+6.84%* | *+2.11%* | *+6.71%* |

Table 7: The effect of feature generation using a newer Wikipedia snapshot (dated March 26, 2006)

## Appendix B. Test Collections for Text Categorization

This Appendix provides detailed description of the test collections we used to evaluate knowledge-based feature generation for text categorization.

### B.1 Reuters-21578

This data set contains one year worth of English-language stories distributed over the Reuters newswire in 1986–1987, and is arguably the most often used test collection in text categorization research. Reuters-21578 is a cleaned version of the earlier release named Reuters-22173, which contained errors and duplicate documents.

The collection contains 21578 documents (hence the name) in SGML format. Of those, 12902 documents are *categorized*, i.e., assigned a category label or marked as not belonging to any category. Other documents do not have an explicit classification; that is, they can reasonably belong to some categories (judged by their content), but are not marked so. Several train/test splits of the collection has been defined, of which ModApte (Modified Apte) is the most commonly used one. The ModApte split divides the collection chronologically, and allocates the first 9603 documents for training, and the rest 3299 documents for testing.

The documents are labeled with 118 categories; there are 0–16 labels per document, with the average of 1.04. The category distribution is extremely skewed: the largest category ("earn") has 3964 positive examples, while 16 categories have only one positive example. Several category sets were defined for this collection:

- 10 largest categories ("earn", "acq", "money-fx", "grain", "crude", "trade", "interest", "ship", "wheat", "corn").

- 90 categories with at least one document in the training set and one in the testing set (Yang, 2001).

- Galavotti, Sebastiani, and Simi (2000) used a set of 115 categories with at least one training example (three categories, "cottonseed", "f-cattle" and "sfr" have no training examples under the ModApte split).

- The full set of 118 categories with at least one positive example either in the training or in the testing set.

Following common practice, we used the ModApte split and two category sets, 10 largest categories and 90 categories with at least one training and testing example.

### B.2 20 Newsgroups (20NG)

The 20 Newsgroups collection (Lang, 1995) is comprised of 19997 postings to 20 Usenet newsgroups. Most documents have a single label, defined as the name of the newsgroup it was sent to; about 4% of documents have been *cross-posted*, and hence have several labels. Each newsgroup contains exactly 1000 positive examples, with the exception of "soc.religion.christian" which contains 997.

Some categories are quite close in scope, for example, "comp.sys.ibm.pc.hardware" and "comp.sys.mac.hardware", or "talk.religion.misc" and "soc.religion.christian". A document posted to a single newsgroup may be reasonably considered appropriate for other groups too (the author may have simply not known of other similar groups, and thus not cross-posted the message); this naturally poses additional difficulty for classification.

It should be noted that Internet news postings are very informal, and therefore the documents frequently contain non-standard and abbreviated words, foreign words, and proper names, as well as a large amount of markup characters (used for attribution of authorship or for message separation).

### B.3 Movie Reviews

The *Movie Reviews* collection (Pang et al., 2002) presents an example of *sentiment classification*, which is different from standard (topical) text categorization. The collection contains 1400 reviews of movies, half of which express positive *sentiment* (opinion) about the movie, and half negative. The reviews were collected from the "rec.arts.movies.reviews" newsgroup, archived at the Internet Movie Database (IMDB, `http://www.imdb.com`). The classification problem in this case is to determine the *semantic orientation* of the document, rather than to relate its content to one of the predefined topics. This problem is arguably more difficult than topical text categorization, since the notion of semantic orientation is quite general. We saw this collection as an opportunity to apply feature generation techniques to this new task.

Recent works on semantic orientation include (Turney & Littman, 2002; Turney, 2002; Pang et al., 2002).[26] The two former studies used unsupervised learning techniques based on latent semantic indexing, estimating semantic distance between a given document and two reference words that represent polar opinions, namely, "excellent" and "poor". The latter work used classical TC techniques.

### B.4 Reuters Corpus Version 1 (RCV1)

RCV1 is the newest corpus released by Reuters (Lewis et al., 2004; Rose, Stevenson, & Whitehead, 2002). It is considerably larger than its predecessor, and contains over 800,000 news items, dated between August 20, 1996 and August 19, 1997. The stories are labeled with 3 category sets, *Topics*, *Industries* and *Regions*.

- *Topics* are most close in nature to the category set of the old Reuters collection (Reuters-21578). There are 103 topic codes, with 3.24 categories per document on the average. The topics are organized in a hierarchy, and the *Hierarchy Policy* required that if a category is assigned to a document, all its ancestors in the hierarchy should be assigned as well. As a result, as many as 36% of all Topic assignments are due to the four most general categories, *CCAT*, *ECAT*, *GCAT*, and *MCAT*. Consequently, the *micro-averaged* performance scores are dominated by these categories (Lewis et al., 2004), and *macro-averaging* becomes of interest.[27] The *Minimum Code Policy* required that each document was assigned at least one Topic and one Region code.

---

26. The field of *genre* classification, which attempts to establish the *genre* of document, is somewhat related to sentiment classification. Examples of possible genres are radio news transcripts and classified advertisements. The work by Dewdney, VanEss-Dykema, and MacMillan (2001) cast this problem as text categorization, using *presentation features* in addition to words. Their presentation features included part of speech tags and verb tenses, as well as mean and variance statistics of sentence and word length, punctuation usage, and the amount of whitespace characters. Using support vector machines for actual classification, the authors found that the performance due to the presentation features alone was at least as good as that achieved with plain words, and that the combined feature set usually resulted in an improvement of several percentage points.

- *Industries* are more fine-grained than Topics, and are therefore harder for classification. These categories are also organized in a hierarchy, although the *Hierarchy Policy* was only partially enforced for them. There are 351,761 documents labeled with Industry codes.

- *Region* codes correspond to geographical places, and are further subdivided into countries, regional groupings and economic groupings. Lewis et al. (2004) argue that Region codes might be more suitable for *named entity recognition* than for text categorization.

In our experiments we used Topic and Industry categories. Due to the sheer size of the collection, processing all the categories in each set would be unreasonably long, allowing to conduct only few experiments. To speed up experimentation, we used a subset of the corpus with 17,808 training documents (dated August 20–27, 1996) and 5341 testing documents (dated August 28–31, 1996). Following the scheme introduced by Brank et al. (2002), we used 16 Topic and 16 Industry categories, which constitute a representative sample of the full groups of 103 and 354 categories, respectively. We also randomly sampled the Topic and Industry categories into 5 sets of 10 categories each. Table 8 gives the full definition of the category sets we used.

As noted by Lewis et al. (2004), the original RCV1 distribution contains a number of errors; in particular, there are documents that do not conform to either *Minimum Code* or *Hierarchy Policy*, or labeled with erratic codes. Lewis et al. (2004) proposed a procedure to correct these errors, and defined a new version of the collection, named *RCV1-v2* (as opposed to the original distribution, referred to as *RCV1-v1*). All our experiments are based on *RCV1-v2*.

## B.5 OHSUMED

OHSUMED (Hersh et al., 1994) is a subset of the MEDLINE database, which contains 348,566 references to documents published in medical journals over the period of 1987–1991. Each reference contains the publication title, and about two-thirds (233,445) also contain an abstract. Each document is labeled with several MeSH categories (MeSH, 2003). There are over 14,000 distinct categories in the collection, with an average of 13 categories per document. OHSUMED is frequently used in information retrieval and text categorization research.

Following Joachims (1998), we used a subset of documents from 1991 that have abstracts, taking the first 10,000 documents for training and the next 10,000 for testing. To limit the number of categories for the experiments, we randomly generated 5 sets of 10 categories each. Table 9 gives the full definition of the category sets we used.

---

27. This is why micro-averaged scores for Topic codes are so much higher than macro-averaged ones, see Section 4.4.2.

| Set name | Categories comprising the set |
|----------|-------------------------------|
| Topic-16 | e142, gobit, e132, c313, e121, godd, ghea, e13, c183, m143, gspo, c13, e21, gpol, m14, c15 |
| Topic-10A | e31, c41, c151, c313, c31, m13, ecat, c14, c331, c33 |
| Topic-10B | m132, c173, g157, gwea, grel, c152, e311, c21, e211, c16 |
| Topic-10C | c34, c13, gtour, c311, g155, gdef, e21, genv, e131, c17 |
| Topic-10D | c23, c411, e13, gdis, c12, c181, gpro, c15, g15, c22 |
| Topic-10E | c172, e513, e12, ghea, c183, gdip, m143, gcrim, e11, gvio |
| Industry-16 | i81402, i79020, i75000, i25700, i83100, i16100, i1300003, i14000, i3302021, i8150206, i0100132, i65600, i3302003, i8150103, i3640010, i9741102 |
| Industry-10A | i47500, i5010022, i3302021, i46000, i42400, i45100, i32000, i81401, i24200, i77002 |
| Industry-10B | i25670, i61000, i81403, i34350, i1610109, i65600, i3302020, i25700, i47510, i9741110 |
| Industry-10C | i25800, i41100, i42800, i16000, i24800, i02000, i34430, i36101, i24300, i83100 |
| Industry-10D | i1610107, i97400, i64800, i0100223, i48300, i81502, i34400, i82000, i42700, i81402 |
| Industry-10E | i33020, i82003, i34100, i66500, i1300014, i34531, i16100, i22450, i22100, i42900 |

Table 8: Definition of RCV1 category sets used in the experiments

# Appendix C. Additional Examples of Feature Generation for Text Categorization

In this Appendix, we list a number of additional feature generation examples.

- **Text:** *'The development of T-cell leukaemia following the otherwise successful treatment of three patients with X-linked severe combined immune deficiency (X-SCID) in gene-therapy trials using haematopoietic stem cells has led to a re-evaluation of this approach. Using a mouse model for gene therapy of X-SCID, we find that the corrective therapeutic gene IL2RG itself can act as a contributor to the genesis of T-cell lymphomas, with one-third of animals being affected. Gene-therapy trials for X-SCID, which have been based on the assumption that IL2RG is minimally oncogenic, may therefore pose some risk to patients.'*

  **Top 10 generated features:** (1) LEUKEMIA; (2) SEVERE COMBINED IMMUNODEFICIENCY; (3) CANCER; (4) NON-HODGKIN LYMPHOMA; (5) AIDS; (6) ICD-10 CHAPTER II: NEOPLASMS; CHAPTER III: DISEASES OF THE BLOOD AND BLOOD-FORMING ORGANS, AND CERTAIN DISORDERS INVOLVING THE IMMUNE MECHANISM; (7) BONE MARROW TRANSPLANT; (8) IMMUNOSUPPRESSIVE DRUG; (9) ACUTE LYMPHOBLASTIC LEUKEMIA; (10) MULTIPLE SCLEROSIS

  **Selected explanations:** (4) a particular cancer type; (6) a disease code of the ICD—International Statistical Classification of Diseases and Related Health Problems

- **Text:** *"Scientific methods in biology"*

| Set name | Categories comprising the set (parentheses contain MeSH identifiers) |
|---|---|
| OHSUMED-10A | B-Lymphocytes (D001402); Metabolism, Inborn Errors (D008661); Creatinine (D003404); Hypersensitivity (D006967); Bone Diseases, Metabolic (D001851); Fungi (D005658); New England (D009511); Biliary Tract (D001659); Forecasting (D005544); Radiation (D011827) |
| OHSUMED-10B | Thymus Gland (D013950); Insurance (D007341); Historical Geographic Locations (D017516); Leukocytes (D007962); Hemodynamics (D006439); Depression (D003863); Clinical Competence (D002983); Anti-Inflammatory Agents, Non-Steroidal (D000894); Cytophotometry (D003592); Hydroxy Acids (D006880) |
| OHSUMED-10C | Endothelium, Vascular (D004730); Contraceptives, Oral, Hormonal (D003278); Acquired Immunodeficiency Syndrome (D000163); Gram-Positive Bacteria (D006094); Diarrhea (D003967); Embolism and Thrombosis (D016769); Health Behavior (D015438); Molecular Probes (D015335); Bone Diseases, Developmental (D001848); Referral and Consultation (D012017) |
| OHSUMED-10D | Antineoplastic and Immunosuppressive Agents (D000973); Receptors, Antigen, T-Cell (D011948); Government (D006076); Arthritis, Rheumatoid (D001172); Animal Structures (D000825); Bandages (D001458); Italy (D007558); Investigative Techniques (D008919); Physical Sciences (D010811); Anthropology (D000883) |
| OHSUMED-10E | HTLV-BLV Infections (D006800); Hemoglobinopathies (D006453); Vulvar Diseases (D014845); Polycyclic Hydrocarbons, Aromatic (D011084); Age Factors (D000367); Philosophy, Medical (D010686); Antigens, CD4 (D015704); Computing Methodologies (D003205); Islets of Langerhans (D007515); Regeneration (D012038) |

Table 9: Definition of OHSUMED category sets used in the experiments

**Top 10 generated features:** (1) Biology; (2) Scientific classification; (3) Science; (4) Chemical biology; (5) Binomial nomenclature; (6) Nature (journal); (7) Social sciences; (8) Philosophy of biology; (9) Scientist; (10) History of biology

**Selected explanations:** (5) the formal method of naming species in biology

- **Text:** *"With quavering voices, parents and grandparents of those killed at the World Trade Center read the names of the victims in a solemn recitation today, marking the third anniversary of the terror attacks. The ceremony is one of many planned in the United States and around the world to honor the memory of the nearly 3,000 victims of 9/11."*

  **Top 10 generated features:** (1) September 11, 2001 attack memorials and services; (2) United Airlines Flight 93; (3) Aftermath of the September 11, 2001 attacks; (4) World Trade Center; (5) September 11, 2001 attacks; (6) Oklahoma City bombing; (7) World Trade Center bombing; (8) Arlington National Cemetery; (9) World Trade Center site; (10) Jewish bereavement

  **Selected explanations:** (2) one of the four flights hijacked on September 11, 2001; (6) a terrorist attack in Oklahoma City in 1995; (8) American military cemetery

- **Text:** *"U.S. intelligence cannot say conclusively that Saddam Hussein has weapons of mass destruction, an information gap that is complicating White House efforts to build support for an attack on Saddam's Iraqi regime. The CIA has advised top administration officials to assume that Iraq has some weapons of mass destruction. But the agency has not given President Bush a "smoking gun," according to U.S. intelligence and administration officials."*

  **Top 10 generated features:** (1) Iraq disarmament crisis; (2) Yellowcake forgery; (3) Senate Report of Pre-War Intelligence on Iraq; (4) Iraq and weapons of mass destruction; (5) Iraq Survey Group; (6) September Dossier; (7) Iraq war; (8) Scott Ritter; (9) Iraq War Rationale; (10) Operation Desert Fox

  **Selected explanations:** (2) falsified intelligence documents about Iraq's alleged attempt to purchase yellowcake uranium; (6) a paper on Iraq's weapons of mass destruction published by the UK government in 2002; (8) UN weapons inspector in Iraq; (10) US and UK joint military campaign in Iraq in 1998

- As another example, consider a pair of contexts that contain the word "jaguar", the first one contains this ambiguous word in the sense of a car model, and the second one—in the sense of an animal.

  - **Text:** *"**Jaguar** car models"*

    **Top 10 generated features:** (1) Jaguar (car); (2) Jaguar (S-Type); (3) Jaguar X-type; (4) Jaguar E-Type; (5) Jaguar XJ; (6) Daimler Motor Company; (7) British Leyland Motor Corporation; (8) Luxury vehicles; (9) V8 engine; (10) Jaguar Racing

    **Top 10 generated features:** (2), (3), (4), (5) — particular Jaguar car models; (6) a car manufacturing company that became a part of Jaguar in 1960; (7)

another vehicle manufacturing company that merged with Jaguar; (9) an internal combustion engine used in some Jaguar car models; (10) a Formula One team used by Jaguar to promote its brand name

– **Text:** *"Jaguar (Panthera onca)"*

**Top 10 generated features:** (1) JAGUAR; (2) FELIDAE; (3) BLACK PANTHER; (4) LEOPARD; (5) PUMA; (6) TIGER; (7) PANTHERA HYBRID; (8) CAVE LION; (9) AMERICAN LION; (10) KINKAJOU

**Top 10 generated features:** (2) a family that include lions, tigers, jaguars, and other related feline species; (10) another carnivore mammal

We also show here a number of examples for generating features using inter-article links.

- **Text:** *"artificial intelligence"*

  **Regular feature generation:** (1) ARTIFICIAL INTELLIGENCE; (2) A.I. (FILM); (3) MIT COMPUTER SCIENCE AND ARTIFICIAL INTELLIGENCE LABORATORY; (4) ARTIFICIAL LIFE; (5) STRONG AI; (6) SWARM INTELLIGENCE; (7) COMPUTER SCIENCE; (8) FRAME PROBLEM; (9) COGNITIVE SCIENCE; (10) CARL HEWITT

  **Features generated using links:** (1) ROBOT; (2) JOHN MCCARTHY (COMPUTER SCIENTIST); (3) ARTIFICIAL CONSCIOUSNESS; (4) MARVIN MINSKY; (5) PLANNER PROGRAMMING LANGUAGE; (6) ACTOR MODEL (a model of concurrent computation formulated by Carl Hewitt and his colleagues); (7) LOGIC; (8) SCIENTIFIC COMMUNITY METAPHOR; (9) NATURAL LANGUAGE PROCESSING; (10) LISP PROGRAMMING LANGUAGE

  **More general features only:** (1) ROBOT; (2) MASSACHUSETTS INSTITUTE OF TECHNOLOGY; (3) PSYCHOLOGY; (4) CONSCIOUSNESS; (5) LISP PROGRAMMING LANGUAGE

- **Text:** *"A group of European-led astronomers has made a photograph of what appears to be a planet orbiting another star. If so, it would be the first confirmed picture of a world beyond our solar system."*

  **Regular feature generation:** (1) PLANET; (2) SOLAR SYSTEM; (3) ASTRONOMY; (4) PLANETARY ORBIT; (5) EXTRASOLAR PLANET; (6) PLUTO; (7) JUPITER; (8) NEPTUNE; (9) MINOR PLANET; (10) MARS

  **Features generated using links:** (1) ASTEROID; (2) EARTH; (3) OORT CLOUD (a postulated cloud of comets); (4) COMET; (5) SUN; (6) SATURN; (7) MOON; (8) MERCURY (PLANET); (9) ASTEROID BELT; (10) ORBITAL PERIOD

  **More general features only:** (1) EARTH; (2) MOON; (3) ASTEROID; (4) SUN; (5) NATIONAL AERONAUTICS AND SPACE ADMINISTRATION

- **Text:** *"Nearly 70 percent of Americans say they are careful about what they eat, and even more say diet is essential to good health, according to a new nationwide health poll in which obesity ranked second among the biggest health concerns."*

  **Regular feature generation:** (1) VEGANISM; (2) VEGETARIANISM; (3) OBESITY; (4) ATKINS NUTRITIONAL APPROACH; (5) BINGE EATING DISORDER; (6) DICK GREGORY; (7) NUTRITION; (8) SUPER SIZE ME; (9) HEALTH INSURANCE; (10) EATING DISORDER

**Selected explanations:** (1) a philosophy of avoiding animal-derived food; (6) an American nutritionist; (7) a documentary film about an individual who eats only McDonald's fast food for one full month.

**Features generated using links:** (1) Raw food diet; (2) Diabetes mellitus; (3) Healthy eating; (4) Body mass index; (5) Omega-3 fatty acid (an important nutritional component); (6) Dieting; (7) Milk; (8) United States[28]; (9) Hypertension; (10) Egg (food)

**More general features only:** (1) United States; (2) Diabetes mellitus; (3) Cancer; (4) Food; (5) McDonald's

## Appendix D. Comparing Knowledge Sources for Feature Generation: Wikipedia versus WordNet

To demonstrate the shortcomings of WordNet as a source for knowledge-based feature generation, we juxtapose WordNet-based and Wikipedia-based feature generation for a sample sentence examined in Section 4.5.1 (we repeat the Wikipedia context classifications for readers' convenience).

We used WordNet version 1.6. In what follows, WordNet synsets are denoted with curly braces, and noun and verb synsets are followed by their immediate hypernym (more general synset), if applicable.

- **Text:** *"Wal-Mart supply chain goes real time"*

- **Most relevant Wikipedia classifications:**

  - Wal-Mart

  - Sam Walton — Wal-Mart founder

  - Sears Holdings Corporation, Target Corporation, Albertsons — prominent competitors of Wal-Mart

  - RFID — Radio Frequency Identification, a technology that Wal-Mart uses very extensively to manage its stock

  - Hypermarket — superstore, a general concept of which Wal-Mart is a specific example

  - United Food and Commercial Workers — a labor union that has been trying to organize Wal-Mart's workers

- **WordNet** :

  - {**Wal-Mart**} → { }; *(word not found in WordNet)*

  - {**supply**} → {indefinite quantity}; {**supply**} → {economic process}; {provision, providing, **supply** supplying} → {activity};

---

28. This classification is quite interesting, as the issue discussed in the input text fragment is very characteristic of the American life style.

- {**supply** provide, render, furnish} → {give}; {issue, **supply**} → {distribute}; {provide, **supply** ply, cater} → {give}; {add, append, **supply**} → {state, say, tell};

- {**chain**, concatenation} → {series}; {**chain**, chemical chain} → {unit, building block}; {**chain**} → {ligament}; {**chain**} → {business, concern, business concern, business organization}; {**chain**} → {restraint, constraint}; {**chain**} → {linear unit}; {range, mountain range, range of mountains, **chain**, mountain chain, chain of maountains} → {geological formation, geology, formation}; {iron, irons, **chain**, chains} → {shackle, bond, hamper, trammel, trammels}; {**chain**, string, strand} → {necklace};

- {**chain**} → {fasten, fix, secure};

- {**go**, spell, tour, turn} → {shift, work shift, duty period}; {crack, fling, **go**, pass, whirl, offer} → {attempt, effort, endeavor, endeavour, try}; {**go**, go game} → {board game};

- {travel, move, **go**, locomote}; {**go**, proceed, move} → {act, move}; *(28 more verb senses omitted for brevity)*

- {real number, **real**}; {**real** (Spanish coin)};

- {**real**, existent}; {**real** (no less than what is stated)}; {**real**, true}; {**real** (not to be taken lightly)}; {**real**, tangible}; {actual, genuine, literal, **real**}; {**real** (economics)}; {substantial, **real**, material}; {**real** (of property)}; {veridical, **real**}; {**real** (founded on practical matters)};

- {very, really, **real**, rattling};

- {**time**, clip} → {case, instance, example}; {**time**} → {time period, period, period of time, amount of time}; {**time**} → {moment, minute, second, instant}; {**time**} → {abstraction}; {clock time, **time**} → {reading, meter reading}; {fourth dimension, **time**} → {dimension}; {**time**} → {experience}; {meter, **time**} → {rhythmicity}; {prison term, sentence, **time**} → {term};

- {clock, **time**} → {quantify, measure}; {**time**} → {schedule}; {**time**} → {determine, shape, influence, regulate}; {**time**} → {adjust, set};

Evidently, WordNet classifications are overly general and diverse because context words cannot be properly disambiguated. Furthermore, owing to lack of proper names, WordNet cannot possibly provide the wealth of information encoded in Wikipedia, which easily overcomes the drawbacks of WordNet. The methodology we proposed does not suffer from the above shortcomings.